\documentclass[10pt,journal,compsoc]{IEEEtran}

\usepackage[utf8]{inputenc} 
\usepackage[T1]{fontenc}    
\usepackage{hyperref}       
\usepackage{url}            
\usepackage{booktabs}       
\usepackage{amsfonts}       
\usepackage{nicefrac}       
\usepackage{microtype}      
\usepackage{physics}
\usepackage{graphicx} 
\usepackage[table]{xcolor}
\usepackage{epstopdf}
\usepackage{wrapfig}
\usepackage{subfigure}
\usepackage{amssymb}
\usepackage{amsmath}
\usepackage{algorithm}
\usepackage{algorithmic}
\usepackage[algo2e,linesnumbered]{algorithm2e}
\usepackage{bm}
\usepackage{amsthm}
\usepackage{bbding}
\usepackage{nicefrac}
\usepackage{enumerate}
\usepackage{enumitem}

\newtheorem{theorem}{Theorem}

\newtheorem*{remark}{Remark}

\newtheorem{definition}{Definition}

\theoremstyle{definition}
\theoremstyle{remark}

\ifCLASSOPTIONcompsoc
  \usepackage[nocompress]{cite}
\else
  \usepackage{cite}
\fi
\ifCLASSINFOpdf
\else
\fi
\begin{document}
%
\title{A Survey of Label-noise Representation Learning: Past, Present and Future}
%
%
%
%

\author{Bo~Han,
        Quanming~Yao,
        Tongliang~Liu,
        Gang~Niu,\\
        Ivor~W.~Tsang,
        James~T.~Kwok,~\IEEEmembership{Fellow,~IEEE}
        and~Masashi~Sugiyama
\IEEEcompsocitemizethanks{\IEEEcompsocthanksitem B. Han is with the Department
of Computer Science, Hong Kong Baptist University, Hong Kong SAR.
E-mail: bhanml@comp.hkbu.edu.hk
\IEEEcompsocthanksitem Q. Yao is with 4Paradigm Inc.
	and Department of Electronic Engineering, Tsinghua University, Beijing China.
E-mail: qyaoaa@connect.ust.hk
\IEEEcompsocthanksitem T. Liu is with the School of Computer Science, The University of Sydney, Sydney, Australia.
E-mail: tongliang.liu@sydney.edu.au
\IEEEcompsocthanksitem G. Niu is with the Imperfect Information Learning Team,
RIKEN AIP, Tokyo, Japan.
E-mail: gang.niu@riken.jp
\IEEEcompsocthanksitem I. W. Tsang is with Australian Artificial Intelligence Institute, University of Technology Sydney, Sydney, Australia.
E-mail: ivor.tsang@uts.edu.au
\IEEEcompsocthanksitem J. T. Kwok is with the Department of Computer Science and Engineering, Hong Kong University of Science and Technology, Hong Kong SAR.\protect\\
E-mail: jamesk@cse.ust.hk
\IEEEcompsocthanksitem M. Sugiyama is with
RIKEN AIP and the Department of Complexity Science and Engineering, The University of Tokyo, Tokyo, Japan.\protect\\
E-mail: sugi@k.u-tokyo.ac.jp}
\thanks{Manuscript received February, 2021.}}

%
%

\markboth{A Survey of Label-noise Representation Learning,~Vol.~XX, No.~XX, February~2021}%
{Shell \MakeLowercase{\textit{et al.}}: Bare Demo of IEEEtran.cls for Computer Society Journals}
%



\IEEEtitleabstractindextext{%
\begin{abstract}
Classical machine learning implicitly assumes that labels of the training data are sampled from a clean distribution, which can be too restrictive for real-world scenarios. However, statistical-learning-based methods may not train deep learning models robustly with these noisy labels.
Therefore, it is urgent to design Label-Noise Representation Learning (LNRL) methods for robustly training deep models with noisy labels. 
To fully understand LNRL, we conduct a survey study. We first clarify a formal definition for LNRL from the perspective of machine learning. 
Then, via the lens of learning theory and empirical study, we figure out why noisy labels affect deep models' performance. 
Based on the theoretical guidance, we categorize different LNRL methods into three directions. 
Under this unified taxonomy, we provide a thorough discussion of the pros and cons of different categories. 
More importantly, we summarize the essential components of robust LNRL, which can spark new directions. 
Lastly, we propose possible research directions within LNRL, such as new datasets, instance-dependent LNRL, and adversarial LNRL. We also envision potential directions beyond LNRL, such as learning with feature-noise, preference-noise, domain-noise, similarity-noise, graph-noise and demonstration-noise. 
\end{abstract}

}

\maketitle

\IEEEdisplaynontitleabstractindextext

%
\IEEEpeerreviewmaketitle

\IEEEraisesectionheading{\section{Introduction}\label{sec:introduction}}

%
%
%
%
\IEEEPARstart{``H}{ow} can a learning algorithm cope with incorrect training examples?'' 
This is the question raised in Dana
Angluin's paper entitled ``Learning From Noisy Examples'' 
in 1988~\cite{angluin1988learning}. 
She made the statement that, ``when the teacher may make independent random errors in classifying the example data, 
the strategy of selecting the most consistent rule for the sample is sufficient, and usually requires a feasibly small number of examples, 
provided noise affects less than half the examples on average''. 
In other words, she claimed that a learning algorithm can cope with incorrect training examples, 
once the noise rate is less than one half under the random noise model. 
Over the last 30 years, her seminal research opened a new door to machine learning, 
since standard machine learning assumes that the label information is fully clean and intact. 
More importantly, her research echoed the real-world environment, as labels or annotations are often noisy and imperfect in real scenarios.

For example, the surge of deep learning comes from 2012, because Geoffrey Hinton's team leveraged AlexNet (i.e., deep neural networks)~\cite{krizhevsky2012imagenet} to win the ImageNet challenge~\cite{deng2009imagenet} with an obvious margin. However, due to the huge quantity of data, the ImageNet-scale dataset was necessarily annotated by distributed workers in Amazon Mechanical Turk~\footnote{\url{https://www.mturk.com/}}. Due to the limited knowledge, distributed workers cannot annotate specific tasks with 100\% accuracy, which naturally brings noisy labels. Another vivid example locates in medical applications, where datasets are typically small. However, it requires domain expertise to label medical data, which often suffers from high inter- and intra-observer variability, leading to noisy labels. We should notice that, noisy labels will cause wrong model predictions, which might further influence decisions that impact human health negatively. Lastly, noisy labels are ubiquitous in speech domains, e.g., Voice-over-Internet-Protocol (VoIP) calls~\cite{reddy2019supervised}. In particular, due to unstable network conditions, VoIP calls are easily prone to various speech impairments, which should involve the user feedback to identify the cause. Such user feedback can be viewed as the cause labels, which are highly noisy, since most of users lack the domain expertise to accurately articulate the impairment in the perceived speech.

All the above noisy cases stem from our daily life, which cannot be avoided. Therefore, it is urgent to build up a robust learning algorithm for handling noisy labels with theoretical guarantees. In this survey paper, we term such a robust learning paradigm \emph{label-noise learning}, and the noisy training data $(x,\bar{y})$ is sampled from a corrupted distribution $p(X,\bar{Y})$, where we assume that the features are intact but the labels are corrupted.
As far as we know, label-noise learning spans over two important ages in machine learning: statistical learning (i.e., shallow learning) and representation learning (i.e., deep learning). In the age of statistical learning, label-noise learning focused on designing noise-tolerant losses or unbiased risk estimators~\cite{natarajan2013learning}. However, in the age of representation learning, label-noise learning has more options to combat with noisy labels, such as designing biased risk estimators or leveraging memorization effects of deep networks~\cite{jiang2018mentornet,han2018co}.



\subsection{Motivation and Contribution}
Label-noise representation learning has become very important for both academia and industry. There are two reasons behind. First, from the essence of the learning paradigm, deep supervised learning requires a lot of well-labeled data, which may require too much cost, especially for many start-ups. However, deep unsupervised learning (even self-supervised learning) is too immature to work very well in complex real-world scenarios. Therefore, as deep weakly-supervised learning, label-noise representation learning naturally has attracted much attention and has become a hot topic. Second, from the aspect of data, many real-world scenarios lack purely clean annotations, such as financial data, web data, and biomedical data. These have directly motivated researchers to explore label-noise representation learning.

As far as we know, there indeed exist three pioneer surveys related to label noise. 
Frenay and Verleysen~\cite{frenay2013classification} focused on discussing label-noise statistical learning, 
instead of label-noise representation learning.
Although Algan et al.~\cite{algan2019image} and Karimi et al.~\cite{karimi2020deep} focused on deep learning with noisy labels, 
both of them only considered image (or medical image) classification tasks. 
Moreover, their surveys were written from the applied perspective, instead of discussing methodology and its beneath theory.
To compensate for them and go beyond, we want to contribute to the label-noise representation learning area as follows.
\footnote{An update-to-date list of papers related to label-noise representation learning is here: \url{https://github.com/bhanML/label-noise-papers}.}
\begin{itemize}[leftmargin=*]
\item From the perspective of machine learning, we give the formal definition for label-noise representation learning (LNRL). 
The definition is not only general enough to include the existing LNRL, but also specific enough to clarify what the goal of LNRL is and how we can solve it.

\item Via the lens of learning theory, 
we provide a deeper understanding why noisy labels affect the performance of deep models. 
Meanwhile, we report the generalization of deep models under noisy labels, which coincides with our theoretical understanding.

\item We perform extensive literature review from the age of representation learning, 
and categorize them in a unified taxonomy in terms of data, objective and optimization. 
The pros and cons of different categories are analyzed. 
We also present a summary of insights for each category.

\item Based on the above observations, we can spark new directions in label-noise representation learning. Beyond label-noise representation learning, we propose several promising future directions, 
such as learning with noisy features, preferences, domains, similarities, graphs, and demonstrations. We hope they can provide some insights.
\end{itemize}

\subsection{Position of the Survey}
The position of this survey is explained as follows.
Frenay and Verleysen~\cite{frenay2013classification} mainly summarized the methods of label-noise statistical learning (LNSL), which cannot be used for deep learning models directly. Note that although both the LNSL and LNRL approaches address the same problem setting, they are fundamentally different. First, the underlying theories should be different due to different hypothesis space (see Section~\ref{sec:thm:obj}); Second, the potential solution should be different due to different models (see Section~\ref{sec:opt}). Meanwhile, LNSL may fail to handle large-scale data with label noise, while LNRL is good at handling such data.

Although Algan et al.~\cite{algan2019image} and Karimi et al.~\cite{karimi2020deep} respectively summarized some methods of label-noise representation learning, both of them discussed from the perspective of applications, i.e., (medical) image analysis. Recently, Song et al.~\cite{song2020learning} summarized some methods of label-noise representation learning from the view of methodology. However, their categorization is totally different from ours in philosophy. In our survey, we first introduce label-noise representation learning from three general views: input data, objective functions and optimization policies, with more theoretical understanding.

\subsection{Organization of the Survey}
The remainder of this survey is organized as follows. Section~\ref{sec:related} provides the related literature of label-noise learning, and the full version can be found in Appendix~1.
Section~\ref{sec:overview} provides an overview of the survey, 
including the formal definition of LNRL, core issues, and a taxonomy of existing works in terms of data, objectives and optimizations.
Section~\ref{sec:data} is for methods that leverage the noise transition matrix to solve LNRL. 
Section~\ref{sec:obj} is for methods that modify the objective function to make LNRL feasible. 
Section~\ref{sec:opt} is for methods that leverage the characteristics of deep networks to address LNRL. 
In Section~\ref{sec:fuworks}, we propose future directions for LNRL. Beyond LNRL, the survey discloses several promising future directions. We conclude this survey in Section~\ref{sec:conclusions}.

\section{Related Literature}
\label{sec:related}
We divide the development of label-noise learning into three stages as follows. Note that the full version of related literature can be found in Appendix 1.
\subsection{Early Stage}
Before delving into label-noise representation learning, 
we give a brief overview of some milestone works in label-noise statistical learning. In 1988, Angluin et al.~\cite{angluin1988learning} proved that a learning algorithm can handle incorrect training examples robustly, when the noise rate is less than one half under the random noise model. Lawrence and Sch\"olkopf~\cite{lawrence2001estimating} constructed a kernel Fisher discriminant to formulate the label-noise problem as a probabilistic model. Bartlett et al.~\cite{bartlett2006convexity} justified that most loss functions are not completely robust to label noise. This means that classifiers based on label-noise learning algorithms are still affected by label noise.

During this period, a lot of works emerged and contributed to this area. For example, Crammer et al.~\cite{crammer2006online} proposed the online Passive-Aggressive perceptron algorithm to cope with label noise. Natarajan et al.~\cite{natarajan2013learning} formally formulated an unbiased risk estimator for binary classification with noisy labels. This work was very important to the area, since it is the first work to provide guarantees for risk minimization under random label noise. Meanwhile, Scott et al.~\cite{scott2013classification} studied the classification problem under the class-conditional noise model, and proposed a way to handle asymmetric label noise. In contrast, van Rooyen et al.~\cite{van2015learning} proposed the unhinge loss to tackle symmetric label noise. Liu and Tao~\cite{liu2015classification} proposed a method using anchor points to estimate the noise rate, and further leveraged importance reweighting to design surrogate loss functions for class-conditional label noise.

In 2015, research of label-noise learning has been shifted from statistical learning to representation learning, since deep learning models have become a mainstream due to its better empirical performance. Therefore, it is urgent to design label-noise representation learning methods for robustly training deep models with noisy labels.

\subsection{Emerging Stage}
There are three seminal works in label-noise representation learning with noisy labels from 2015. For example, Sukhbaatar et al.~\cite{sukhbaatar2014training} introduced an extra but constrained linear ``noise'' layer on top of the softmax layer, which adapts the network outputs to model the noisy label distribution. Reed et al.~\cite{reed2014training} augmented the prediction objective with the notion of consistency via soft and hard bootstrapping. Intuitively, this bootstrapping procedure provides the learner to disagree with an inconsistent training label, and re-label the training data to improve its label quality. Azadi et al.~\cite{azadi2015auxiliary} proposed an auxiliary image regularization technique, which exploits the mutual context information among training data, and encourages the model to select reliable labels.

Following the seminal works, Goldberger et al.~\cite{goldberger2016training} introduced a nonlinear ``noise'' adaptation layer on top of the softmax layer. Patrini et al.~\cite{patrini2017making} proposed the forward and backward loss correction approaches simultaneously. Both Wang et al.~\cite{wang2017robust} and Ren et al.~\cite{ren2018learning} leveraged the same philosophy, namely data reweighting, to learn with label noise. Jiang et al.~\cite{jiang2018mentornet} is the first to leverage small-loss tricks to handle label noise. However, they trained only a single network iteratively, which inherits the accumulated error. To alleviate this, Han et al.~\cite{han2018co} trained two deep neural networks, and each network backpropagated the data selected by its peer network and updated itself.

In the context of representation learning, classical methods, such as estimating the noise transition matrix, regularization and designing losses, are still prosperous for handling label noise. For instance, Hendrycks et al.~\cite{hendrycks2018using} leveraged trusted examples to estimate the gold transition matrix, which approximates the true transition matrix well. Han et al.~\cite{han2018masking} proposed a ``human-in-the-loop'' idea to easily estimate the transition matrix. Zhang et al.~\cite{zhang2017mixup} introduced an implicit regularization called mixup, which constructs virtual training data by linear interpolations of features and labels in training data. Zhang et al.~\cite{zhang2018generalized} generalized both the categorical cross entropy loss and mean absoulte error loss by the negative Box-Cox transformation. Ma et al.~\cite{ma2018dimensionality} developed a dimensionality-driven learning strategy, which can learn robust low-dimensional subspaces capturing the true data distribution.

\subsection{Flourished Stage}
\label{sec:intro:mat}

Since 2019, label-noise representation learning has become flourished in the top conference venues. Arazo et al.~\cite{arazo2019unsupervised} formulated clean and noisy samples as a two-component (clean-noisy) beta mixture model on the loss values. Hendrycks et al.~\cite{hendrycks2019using} empirically demonstrated that pre-training can improve model robustness against label corruption for large-scale noisy datasets. Under the criterion of balanced error rate (BER) minimization, Charoenphakdee et al.~\cite{charoenphakdee2019symmetric} proposed the barrier hinge loss. In contrast to selected samples via small-loss tricks, Thulasidasan et al.~\cite{thulasidasan2019combating} introduced the abstention-based training, which allows deep networks to abstain from learning on confusing samples but to learn on non-confusing samples. Following the re-weighting strategy, Shu et al.~\cite{shu2019meta} parameterized the weighting function adaptively as a one-layer multilayer perceptron called Meta-Weight-Net.

Menon et al.~\cite{menon2019can} mitigated the effects of
label noise from an optimization lens, which naturally introduced the partially Huberised loss. Nguyen et al.~\cite{nguyen2019self} proposed a self-ensemble label filtering method to progressively filter out the wrong labels during training. Li et al.~\cite{li2020dividemix} modeled the per-sample loss distribution with a mixture model to dynamically divide the training data into a labeled set with clean samples and an unlabeled set with noisy samples. Lyu et al.~\cite{lyu2019curriculum} proposed a provable curriculum loss, which can adaptively select samples for robust stagewise training. Han et al.~\cite{han2020sigua} proposed a versatile approach called scaled stochastic integrated gradient underweighted ascent (SIGUA). SIGUA uses stochastic gradient decent on good data, while using scaled stochastic gradient ascent on bad data rather than dropping those data. 5 years after the birth of \textit{Clothing1M}, Jiang et al.~\cite{jiang2020cont} proposed a new but realistic type of noisy dataset called ``web-label noise'' (or \textit{red noise}).

\section{Overview}\label{sec:overview}
In this section, we first provide the notation used throughout the paper in Section~\ref{sec:notation}. 
A formal definition of the LNRL problem is given in Section~\ref{sec:pro-def} with concrete examples. As the LNRL problem relates to many machine learning problems, we discuss their relatedness and difference in Section~\ref{sec:rl-pro}. In Section~\ref{sec:core-issue}, we reveal the core issues that make the LNRL problem hard. Then, according to how existing works handle the core issues, we present a unified taxonomy in Section~\ref{sec:theorem}.

\subsection{Notation}
\label{sec:notation}

Let $x$ be features and $y$ be labels. Consider a supervised learning task $T$: LNRL deals with a data set $\mathcal{D} = \{\bar{\mathcal{D}}^{\text{tr}},\mathcal{D}^{\text{te}}\}$ consisting of training set $\bar{\mathcal{D}}^{\text{tr}} = \{(x_i,\bar{y}_i)\}_{i=1}^N$ and test set $\mathcal{D}^{\text{te}} = \{x^{\text{te}}\}$, 
where training set $\bar{\mathcal{D}}^{\text{tr}}= \{(x_i,\bar{y}_i)\}_{i=1}^N$ is independently drawn from a corrupted distribution $\bar{D} = p(X,\bar{Y})$ ($\bar{Y}$ denotes the corrupted label).
Note that ($X,Y$) denotes the variable, while ($x,y$) denotes its sampled value.
For the corrupted distribution $p(X,\bar{Y})$, we assume that the features are intact but the labels are corrupted. Let $p(X, Y)$ be the ground-truth (i.e., non-corrupted) joint probability distribution of features $x$ and label $y$, 
and $f^*$ be the (Bayes) optimal hypothesis from $x$ to $y$. 
To approximate $f^*$, 
the objective requires a hypothesis space $\mathcal{H}$ of hypotheses 
$f_{\theta}(\cdot)$ parameterized by $\theta$. 
An algorithm contains the
optimization policy to search through $\mathcal{H}$ in order to find $\theta^*$ that corresponds to the optimal function in the hypothesis for $\bar{\mathcal{D}}^{\text{tr}}$: $f_{\theta^*} \in \mathcal{H}$.

Intuitively, LNRL learns to discover $f_{\theta^*}$ by fitting $\bar{\mathcal{D}}^{\text{tr}}$ robustly, which can assign correct labels for $\mathcal{D}^{\text{te}}$. LNRL methods robustly train deep neural networks with noisy labels, 
where hypotheses $f_{\theta}(\cdot)$ can be modeled by deep neural networks. Since the hypothesis space $\mathcal{H}$ is sufficiently complex for deep neural networks, $f_{\theta^*} \in \mathcal{H}$ is expected to approximate the Bayes optimal $f^*$ well~\cite{raghu2017expressive}. 

\begin{table*}[!tp]
\centering
\caption{Three LNRL examples based on Definition 2.2.}
\vspace{-10px}
\begin{tabular}{c|c|c}
	\hline
	$T$ & $E = (x,y)$ & $P$ \\\hline
	web-scale image classification  &  (ImageNet, crowdsourced labels) &  test accuracy \\\hline
	intelligent healthcare &  (medical data, annotations by variability) &  error rate \\\hline
	service call analysis &  (perceived speech, user rating) &  quality rate of call \\\hline
\end{tabular}
\label{tab:LNRL-exp}
\end{table*}

\subsection{Problem Definition}
\label{sec:pro-def}
As LNRL is naturally a sub-area in machine learning, 
before giving the definition of LNRL, let us recall how machine learning is defined in literature.
We borrow Tom Mitchell's definition here, 
which is shown in Definition~\ref{def:clsml}.

\begin{definition}
\label{def:clsml}
(Machine Learning~\cite{mitchell1997machine,mohri2018foundations}).
 A computer program is said to learn from experience $E$ with respect to some classes of task $T$ and performance measure $P$ if its performance can improve with $E$ on $T$ measured by $P$.
\end{definition}

The above definition is quite classical, which has been widely adopted in the machine learning community.
It means that a machine learning problem is defined by three key components: $E$, $T$ and $P$. 
For instance, consider a speech recognition task ($T$, e.g., 
Apple Siri\footnote{\url{https://en.wikipedia.org/wiki/Siri}}), 
machine learning programs can improve its recognition accuracy ($P$) via training with a large-scale speech data set ($E$) offline.
Another example  of $T$ is the hot topic in the security area, 
called empirical defense~\cite{Goodfellow2015ExplainingAH}. 
In a high-level, 
machine learning algorithms can make deep neural networks defensive against malicious cases. 
Specifically, a stop sign crafted by malicious people may cause an accident to autonomous vehicles, 
which employ deep neural networks to recognize the sign. 
However, 
after adversarial training with adversarial examples ($E$), 
the robust generalization ($P$) of deep neural networks can improve a lot, 
which may avoid the above accident with large probability. 

The above-mentioned classical applications of machine learning require a lot of ``correctly'' supervised information $\{(x^{(i)},y^{(i)})\}_{i=1}^N$ for the given tasks. 
However, this may be difficult and even impossible. 
As far as we know, LNRL is a special case of machine learning, which belongs to weakly supervised learning~\cite{zhou2018brief}. 
Intuitively, LNRL exactly targets at acquiring good learning performance with ``incorrectly'' (a.k.a., noisy) supervised information provided by data set $\bar{D}$, 
drawn independently from a corrupted distribution $p(X,\bar{Y})$.
The noisy supervised information refers to training data set $\bar{D}^{\text{tr}}$, which consists of the intact input features $x^{(i)}$ but with corrupted labels $\bar{y}^{(i)}$. More important, LNRL focuses on training deep neural networks robustly, which has many special characteristics, such as memorization effects~\cite{arpit2017closer}. Formally, we define LNRL in Definition~\ref{LNRL}.

\begin{definition}\label{LNRL}
(Label-Noise Representation Learning (LNRL)). 
LNRL is a special but common case of machine learning problems (specified by $E$, $T$ and $P$), 
where $E$ contains noisy supervised information for the target $T$. 
Meanwhile, deep neural networks will be leveraged to model the target $T$ directly. 
\end{definition}

To understand this definition better, let us show three classical scenarios of LNRL (Table \ref{tab:LNRL-exp}):
\begin{itemize}[leftmargin=*]
\item \textit{Image}: Large-scale image data (e.g., ImageNet~\cite{Russakovsky2015ImageNetLS}) 
is the key factor to drive the second surge of deep learning from 2012. Note that it is impossible to annotate such large-scale data individually, which motivates us to leverage crowdsourcing technique (e.g., Amazon Mechanical Turk). However, the quality of crowdsourced data is normally low with a certain degree of label noise. Therefore, an important task ($T$) is to robustly training deep neural networks with crowdsourced data ($E$), and the trained deep models can be evaluated via the test accuracy ($P$).

\item \textit{Healthcare}: Healthcare is highly related to each individual, whose data requires machine learning technique to analyze deeply and intelligently. However, intelligent healthcare ($T$) requires domain expertise to label medical data first, which often suffers from high inter- and intra-observer variability, leading to noisy medical data ($E$). We should notice that, noisy labels will cause a high error rate ($P$) of deep model predictions, which might further influence decisions that impact human health negatively.

\item \textit{Speech}: In the speech recognition task (e.g., Apple Siri), the machine learning program can improve its recognition accuracy via training with a large-scale speech data set offline. However, noisy labels are ubiquitous in speech domains, e.g., the task of rating for service calls ($T$). Due to the difference of the personal mood and understanding, service calls are easily prone to different rating ($E$) for the same service. Such rating can be viewed as labels, which are highly noisy since most of users lack the domain expertise to accurately rate the speech service. Therefore, it is critical to robustly train deep neural networks with the user rating ($E$), and evaluate the trained model via the quality rate of service calls ($P$).
\end{itemize}

As noisy supervised information related to $T$ is directly contained in $E$, it is quite natural that common deep supervised learning approaches will fail on LNRL problems. One of the recent findings (i.e., memorization effects~\cite{arpit2017closer}) in the deep learning area may explain this: due to the high model capacity, deep neural networks will eventually fit and memorize label noise. Therefore, when facing the noisy data $E$, LNRL methods make the learning of the target $T$ feasible by leveraging the intrinsic characteristics of deep neural networks, e.g., memorization effects.

\subsection{Relevant Learning Problems}
\label{sec:rl-pro}
In this section, we discuss the relevant learning problems of LNRL. The relatedness and difference with respect to LNRL are clarified as follows.
\begin{itemize}[leftmargin=*]
\item \textit{Semi-supervised Learning (SSL)}~\cite{zhu2009introduction,tarvainen2017mean,miyato2018virtual,berthelot2019mixmatch,zhang2017mixup} learns the hypothesis $f$ from experience $E$ consisting of both labeled and unlabeled data, where unlabeled data will be normally given pseudo labels. Since the labeling process may be not fully correct and noisy, SSL has some relation with LNRL. However, standard SSL methods assumes that labeled data are fully clean, which is different from LNRL, where labeled data are still noisy to some degree.
\item \textit{Positive-unlabeled Learning (PUL)}~\cite{elkan2008learning,kiryo2017positive,hsieh2019classification,ishida2018binary} learns the hypothesis $f$ from experience $E$ consisting of only positive labeled and unlabeled data. Similar to SSL, unlabeled data will be normally given pseudo labels. However, PUL assumes that labeled data are fully clean and only positive.
\item \textit{Complementary Learning (CL)}~\cite{ishida2017learning,yu2018learning,ishida2019complementary,feng2019learning} specifies a class that a pattern does NOT belong to. Namely, CL learns the hypothesis $f$ from experience $E$ consisting of only complementary data. Since the labeling process cannot fully exclude the uncertainty, namely belonging to which categories, CL has some relation with LNRL. However, CL requires that all diagonal entries of the transition matrix are zeros. Sometimes, the transition matrix is not required to be invertible empirically. 
\item \textit{Unlabeled-unlabeled Learning (UUL)}~\cite{du2013clustering,lu2018minimal,lu2020mitigating} 
allows training a binary classifier from two unlabeled datasets with different class priors. 
Different from SSL/PUL, there are two sets of unlabeled data in UUL instead of one set.
\end{itemize}

\subsection{Empirical Risk under Label Noise}
\label{sec:core-issue}
When machine learning is carried out in an ideal environment, the data should be with clean supervision. 
In this survey, we consider the classification problem.
Then, the $\ell$-risk should be as follows 
\begin{equation}
\label{eq:clean-risk}
R_{\ell,D}(f_{\theta}) := \mathbb{E}_{(x,y)\sim D} [\ell(f_{\theta}(x), y)],
\end{equation}
where $(x,y)$ is the clean example drawn independently from clean distribution $D$, $f_{\theta}$ is a learning model 
(e.g., a deep neural network) parameterized by $\theta$ and $\ell$ is normally the cross-entropy loss.

However, when machine learning is used in the real-world environment,
labels can be corrupted. 
Namely, the $\ell$-risk under the noisy distribution should be
$\mathbb{E}_{(x,\bar{y})\sim \bar{D}} [\ell(f_{\theta}(x), \bar{y})]$,
where $(x, \bar{y})$ is the sample (with possibly corrupted label) drawn independently from noisy distribution $\bar{D}$. 
Under finite samples, 
the empirical $\tilde{\ell}$-risk under $\bar{D}$ should be as follows.
\begin{equation}
\label{eq:noisy-risk}
\widehat{R}_{\tilde{\ell},\bar{D}}(f_{\theta})
:= \nicefrac{1}{N}\sum\nolimits_{i=1}^N \tilde{\ell}(f_{\theta}(x_i), \bar{y}_i),
\end{equation}
where $\tilde{\ell}$ is a suitably modified loss, which also serves as a noise-tolerant objective.

\begin{figure}
\begin{center}
\centerline{\includegraphics[width=0.35\textwidth]{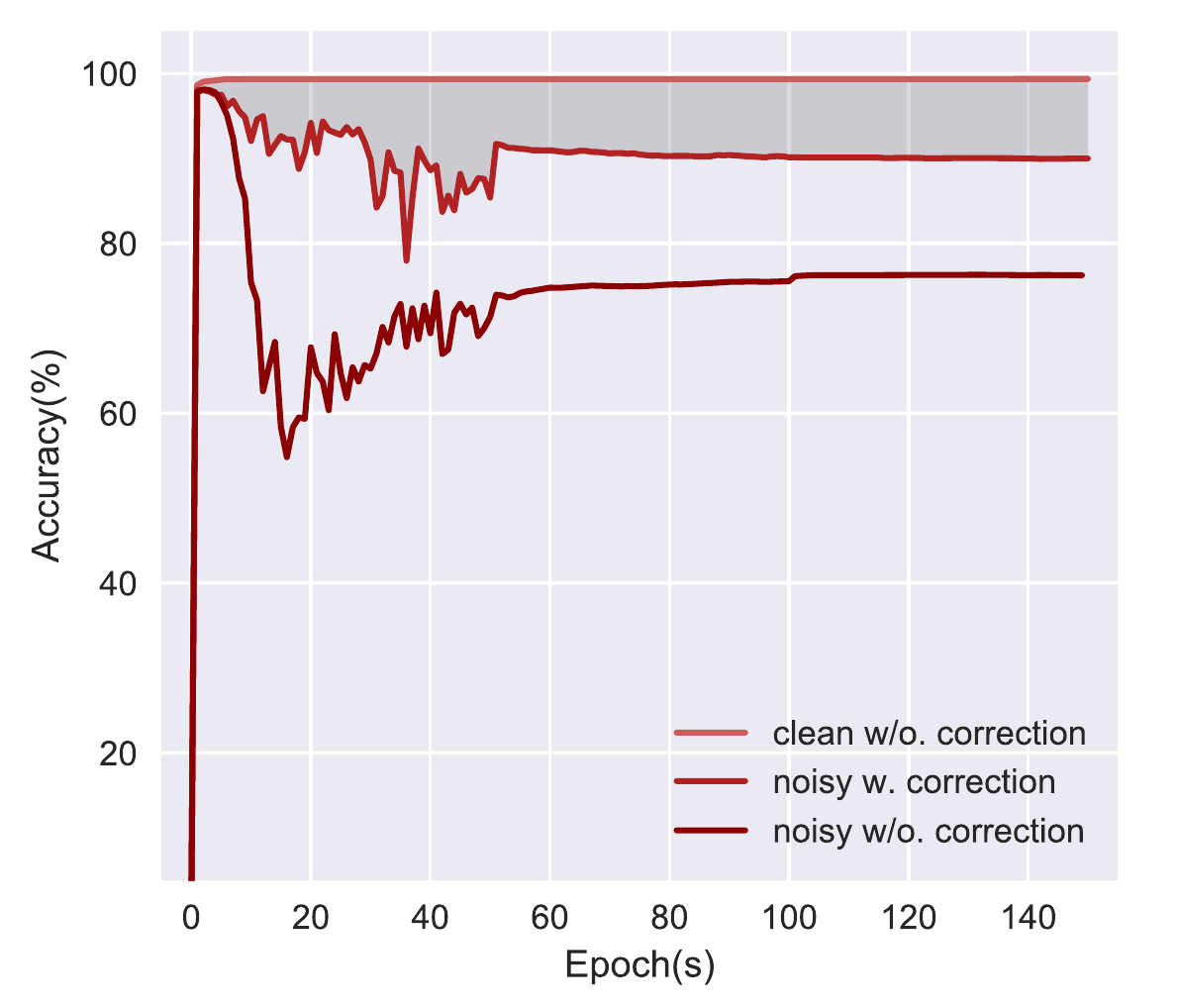}}
\vspace{-10px}
\caption{We empirically demonstrate the generalization difference between original $\ell$ and corrected $\tilde{\ell}$ (cf. Theorem~\ref{fw-theorem} in Section~\ref{sec:bfcorr}). 
	We choose \textit{MNIST} with 35\% of uniform noise as noisy data. 
	There is an obvious gap between $\ell$ and $\tilde{\ell}$ on noisy \textit{MNIST}.}
\label{motivation-fig}
\end{center}
\end{figure}

Here, we empirically demonstrate the generalization difference between $\ell$ and $\tilde{\ell}$ under label noise (Figure~\ref{motivation-fig}), where $\tilde{\ell}$ is a forward-corrected loss (Theorem~\ref{fw-theorem} in Section~\ref{sec:bfcorr}) and experimental details can be found in Appendix 3. 
Generally, the aim of LNRL is to ``construct'' such noise-tolerant $\tilde{\ell}$ that the learned $f_{\theta}$ by minimizing \eqref{eq:noisy-risk} approximates the optimal $f_{\theta^*}$ that minimizes \eqref{eq:clean-risk} well. Specifically, via a suitably constructed $\tilde{\ell}$, we can learn a robust deep classifier $f_{\theta}$ from the noisy training examples that can assign clean labels for test instances. 
Next, 
we first take a theoretical look at label-noise learning, which will help us build $\tilde{\ell}$ more effectively.

\subsection{Theoretical Understanding}
\label{sec:theorem}

In contrast to \cite{algan2019image,karimi2020deep}, via the lens of learning theory, 
we provide a systematic way to understand LNRL. 
Our focus is to explore why noisy labels affect the performance of deep models. To figure it out, we should rethink the essence of learning with noisy labels. Normally, there are three key ingredients in label-noise learning problems, including the data, the objective function and the optimization policy.

In a high-level, 
there are three rules of thumb, which explain how to handle LNRL.
\begin{itemize}[leftmargin=*]
\item For \textit{data}, 
the key is to discover the underlying noise transition pattern, which directly links the clean class posterior and the noisy class posterior. Based on this insight, it is critical to design an accurate estimator of noise transition matrix $T$.

\item For the \textit{objective function}, 
the key is to design noise-tolerant $\tilde{\ell}$ in \eqref{eq:noisy-risk}, which can be more robust than standard loss functions. Based on this insight, it is critical to learn a robust classifier on noisy data, which can provably converge to the learned classifier on clean data.

\item For the \textit{optimization policy}, 
the key is to explore the dynamic process of optimization, which relates to memorization. Based on this insight, it is critical to trade-off overfit/underfit in training deep networks, such as early stopping and small-loss tricks, where small-loss tricks backpropogate small-loss data based on memorization effects of deep networks.
\end{itemize}

\begin{figure*}
	\centering
	\includegraphics[width=0.65\textwidth]{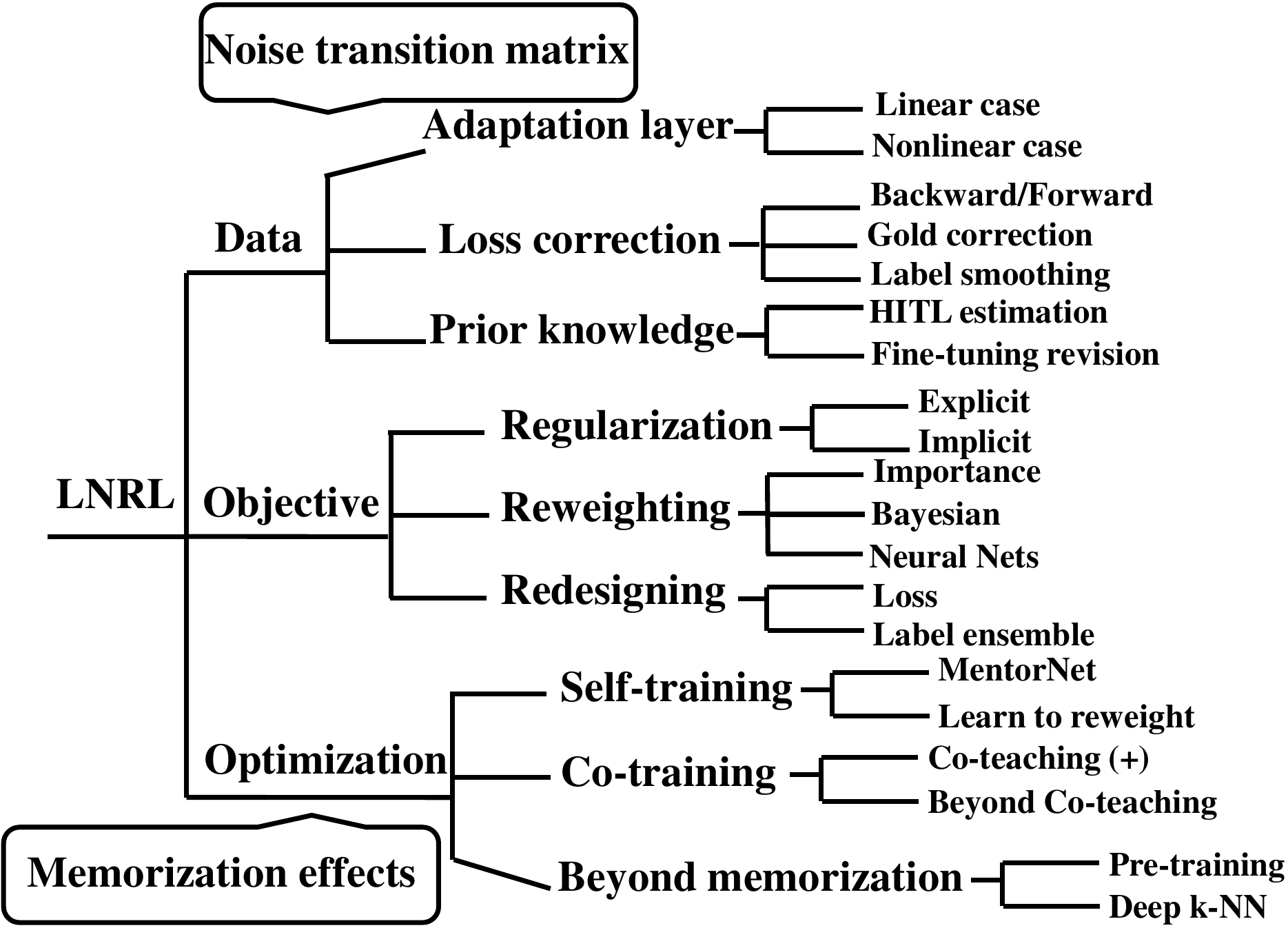}
	\caption{A taxonomy of LNRL based on the focus of each method. For each technique branch, we list a few representative works here.}
	\label{fig:taxonomy}
\end{figure*}

\subsubsection{Perspective of Data}
\label{sec:thm:idata}

Specifically, 
from the perspective of the data, the focus is to build up the noise transition matrix, 
which models the process of label corruption. 
In general, there are two types of label noise: 
instance-dependent label noise (e.g., $p(\bar{Y}|Y,X)$) \cite{menon2018learning} and instance-independent label noise (e.g., $p(\bar{Y}|Y)$) \cite{patrini2017making}. 
For instance-dependent label noise, the noise transition matrix can be represented as 
$T(X)$, 
which depends on features. 
However, 
it can be ill-posed to learn the transition matrix $T(X)$ by only exploiting noisy data, 
i.e., the transition matrix is unidentiable~\cite{menon2018learning,cheng2020learning}.
Therefore, 
we emphasize instance-independent label noise here, 
and the noise transition matrix can be represented as $T$, 
which is independent of features. 

In this case, the noise transition matrix $T$ approximately models the process of label corruption. 
An instance $x^i$ is said to be an anchor point of the $i$-th clean class if $p(Y = e_i|x^i)=1$, where $Y = e_i$ means $Y$ belongs to the $i$-th class. The transition matrix can be obtained via
\begin{align}
p(\bar{Y} = e_j | x^i) 
& = \sum\nolimits_{k=1}^{C} p(\bar{Y} = e_j | Y = e_k, x^i) p(Y = e_k|x^i),\notag
\\
& = p(\bar{Y} = e_j | Y = e_i, x^i)p(Y = e_i|x^i),\notag
\\
& = p(\bar{Y} = e_j | Y = e_i, x^i) = T_{ij}.\label{eq:trans}
\end{align}
Note that if 
anchor points are hard to identify, we can use $x^i = \arg\max_{x}p(\bar{Y} =i|x)$ ~\cite{liu2015classification}. This transition matrix is very important, since it can bridge the noisy class posterior and clean class posterior, i.e., $p(Y|x) = T^{-1}p(\bar{Y}|x)$. 
In practice, this transition matrix has been employed to build a risk-consistent estimator via loss correction or a classifier-consistent estimator via hypotheses correction~\cite{xia2019anchor}. 
Besides, for inconsistent algorithms, the diagonal entries of this matrix are used to select reliable examples for further robust training~\cite{han2018co}.

\subsubsection{Perspective of Objective Function}
\label{sec:thm:obj}

From the perspective of the objective function, 
the focus is to derive the statistical consistency 
guarantees for robust 
$\tilde{\ell}$~\cite{natarajan2013learning,patrini2016loss,menon2019can}.
Let $f$ be a deep network with $d$ layers and ReLU active function,
$R^*=R_D(f^*)$ denote the Bayes risk for Bayes optimal classifier $f^*$ under the clean distribution $D$~\cite{mohri2018foundations},
and $\hat{f} = \arg\min_{f\in\mathcal{H}} \widehat{R}_{\tilde{\ell},\bar{D}}(f)$
where $L_{\rho}$ is the Lipschitz constant of $\tilde{\ell}$.
Assume the Frobenius norm of the weight matrices $W_1,\ldots,W_d$ are at most $M_1,\ldots,M_d$
and $x$ be upper-bounded by $B$, i.e., $\Vert x \Vert \leq B$ for any $x$. 
With probability at least $1-\delta$, if $\ell$ is classification-calibrated~\cite{van2015learning}, 
there exists a non-decreasing function $\xi_{\ell}$ with $\xi_{\ell}(0)=0$ such that 
\begin{equation*}
\begin{split}
& R_D (\hat{f})  - R^* \leq \xi_{\ell}\big(\min\nolimits_{f\in\mathcal{H}} R_{\ell,D}(f) - \min\nolimits_f R_{\ell,D}(f) 
\\
& \quad + 4L_{\rho}\mathcal{R}(\mathcal{H}) + 2 \sqrt{\log(\nicefrac{1}{\delta}) / 2N}\big), 
\\
& \! \leq \! \xi_{\ell}\big(\min_{f\in\mathcal{H}} R_{\ell,D}(f) 
\! - \! \min_f R_{\ell,D}(f) 
\! + \! 4L_{\rho} C 
\! + \! 2 \sqrt{\log(\nicefrac{1}{\delta})/2N}\big),
\end{split}
\end{equation*}
where 
$C = B(\sqrt{2 d \log 2} \! + \! 1) \prod\nolimits_{i=1}^d M_i/\sqrt{N}$ and
$R_D(\hat{f}) = \mathbb{E}_{(x,y) \sim D}[1_{\{\text{sign}(\hat{f}(x))\neq y\}}]$ 
denotes the risk of $\hat{f}$ w.r.t. the 0-1 loss.
Note that
for a deep neural network, 
its Rademacher complexity $\mathcal{R}(\mathcal{H})$ of the function class $\mathcal{H}$ is upper-bounded by $C$~\cite{golowich2018size}.

\begin{remark}
The above conclusions denote that the learned $\hat{f}$ (using noise-tolerant $\tilde{\ell}$ over noisy $\bar{D}$) can approach the Bayes optimal $f^*$, when increasing the richness of the class $\mathcal{H}$ and the data size~$N$.
\end{remark}

Note that $\min_{f\in\mathcal{H}} R_{\ell,D}(f) - \min_f R_{\ell,D}(f)$ denotes the approximation error for employing the hypothesis class $\mathcal{H}$.
According to the universal approximation theorem~\cite{anthony2009neural}, 
if a certain deep network model is employed, $\mathcal{H}$ will be a universal hypothesis class and thus contains the Bayes optimal classifier. 
Then, $\min_{f\in\mathcal{H}} R_{\ell,D}(f) - \min_f R_{\ell,D}(f)=0$. 
This means that by employing a proper deep network, 
the upper bound will converge to zero by increasing the training sample size $N$. Since $R_D(\hat{f})$ is always bigger than or equal to $R^*$, 
we have that $R_D(\hat{f})$ will converge to $R^*$. 
This further means that $\hat{f}$ learned from noisy data (independently drawn from $\bar{D}$) will converge to Bayes optimal $f^*$ defined by the clean data.


\subsubsection{Perspective of Optimization Policy}
From the perspective of the optimization policy,
the focus is to explore the dynamic process of optimization.
Take the early stopping~\cite{li2020gradient}, 
which is a simple yet effective trick to avoid overfitting on noisy labels,
as an example. 
Assume an 
initial weight matrix having entries following the standard normal distribution, 
namely $W^0 \sim \mathcal{N}(0,1)$ entries, and $W^{\tau}$ is updated via stochastic gradient descent with step size $\eta$, 
i.e., $W^{\tau+1} = W^{\tau} - \eta \nabla \ell(W^{\tau})$. If $\varepsilon_0 \leq \delta\lambda(C)^2/K^2$ and $\rho \leq \delta/8$ (cf. Definition 1.2 in \cite{li2020gradient}), 
then after $I \propto \nicefrac{\lVert C \rVert^2}{\lambda(C)}$ steps, there are two conclusions \cite{li2020gradient} with high probability. First, the model $W^I$ predicts the true label function $\hat{y}(x)$ for all input $x$ that lies within the $\varepsilon_0$-neighborhood of a cluster center $\{c_k\}_{k=1}^K$.
Namely, $ \hat{y}(x) = \arg\min_{l}|f_{W^I}(x) - \alpha_l|$, 
where $\{\alpha_l\}_{l=1}^{\bar{K}}\in[-1,1]$ denotes the labels associated with each class, and each label belongs to $\bar{K} (\leq K$) classes (cf. Definition 1.1 in \cite{li2020gradient}). Second, for all training samples, the distance to the initial weight matrix satisfies
\begin{equation*}
\lVert W^{\tau} - W^0 \rVert_\mathrm{F} \lesssim \big(\sqrt{K} + \nicefrac{\tau \varepsilon_0 K^2}{\lVert C \rVert^2} \big),
\end{equation*}
where $0 \leq \tau \leq I$ and $A \lesssim B$ denotes $A \leq \beta B$ with some constant $\beta$. 
\begin{remark}
The above conclusions demonstrate that gradient descent with early stopping (i.e., $I$ steps) 
can be robust when training deep neural networks. 
Moreover, the final network weights do not stray far from the initial weights for robustness, 
since the distance between the initial model and final model grows with the square root of the number of clusters $\sqrt{K}$.
\end{remark}

Intuitively, due to memorization effects, deep neural networks will eventually overfit noisy training data~\cite{zhang2016understanding,arpit2017closer}. Thus, it is a good strategy to stop training early, when deep neural networks fit clean training data in first few epochs. This denotes the robust weights are not far away from the initial weights.



\subsection{Taxonomy}
\label{sec:taxonomy}

Based on the above theoretical understanding, we categorize these works into three general perspectives:
\begin{enumerate}[leftmargin=*]
\item \textit{Data}~\cite{van2017theory}: 
From the perspective of data, 
the key is to build up the noise transition matrix $T$, which explores the data relationship between clean and noisy labels. Take loss correction as an example. We first model and estimate $T$ between latent $Y$ and observed $\bar{Y}$. Then, via the estimated matrix, different correction techniques can generate $\tilde{\ell}$ from the original $\ell$. 
Mathematically, 
$\widehat{R}_{\tilde{\ell},\bar{D}}(f_{\theta}) := \frac{1}{n}\sum_{i=1}^n \tilde{\ell}(f_{\theta}(x_i), \bar{y}_i)$,
where 
$\ell \xrightarrow{T} \tilde{\ell}$, and $\tilde{\ell}$ is a corrected loss transitioning from $\ell$ via $T$.

\item \textit{Objective}: From the perspective of the objective, we can construct $\tilde{\ell}$ by augmenting the objective function, 
i.e., the original $\ell$, 
via either explicit or implicit regularization.
For instance, we may augment $\ell$ by an auxiliary regularizer explicitly. Meanwhile, we may augment $\ell$ by designing implicit regularization algorithms, such as soft-/hard-bootstrapping~\cite{reed2014training}
and virtual adversarial training (VAT)~\cite{miyato2018virtual}. 
Meanwhile, we can construct $\tilde{\ell}$ by reweighting the objective function $\ell$. Lastly, we can also construct and design $\tilde{\ell}$ directly. Thus, $\tilde{\ell}$ has three options:
\begin{itemize}
\item $\tilde{\ell} = \ell + r$, where $r$ denotes a regularization; 

\item $\tilde{\ell} = \sum_i w_i\ell_i$, where $\ell_i$ denotes $i$-th sub-objective 
with the coefficient $w_i$; 

\item $\tilde{\ell}$ has a special format $\ell'$ independent of $\ell$.
\end{itemize}

\item \textit{Optimization}~\cite{arpit2017closer,zhang2016understanding}: From the perspective of optimization, we can construct $\tilde{\ell}$ by leveraging the memorization effects of deep models. For example, due to the memorization effects, deep models tend to fit easy (clean) patterns first, and then over-fit complex (noisy) patterns gradually. Based on this observation, we can backpropagate the small-loss examples, which is equal to constructing the restricted $\tilde{\ell}$
where $\tilde{\ell} = \text{sort}(\ell,1-\tau)$, namely, sorting $\ell$ from small to large, and fetching $1-\tau$ percentage of small losses ($\tau$ is the noise rate).
\end{enumerate}

Accordingly, existing works can be categorized into a unified taxonomy as shown in Figure~\ref{fig:taxonomy}. 
We will detail each category in the sequel. Note that the discussion of three perspectives can be found in Appendix 2.




\section{Data}\label{sec:data}
Methods in this section solve the LNRL problem by estimating the noise transition matrix, which builds up the relationship between latent clean labels and observed noisy labels. First, we explain what the noise transition matrix is and why this matrix is important. Then, we introduce three common ways to leverage the noise transition matrix for combating label noise. The first way is to leverage an adaptation layer in the end-to-end deep learning system to mimic the noise transition matrix, which bridges latent clean labels and observed noisy labels. The second way is to estimate the noise transition matrix empirically, and further correct the cross-entropy loss by the estimated matrix. Lastly, the third way is to leverage prior knowledge to ease the estimation burden.

\subsection{Noise Transition Matrix}
Before introducing three common ways, we first define what the noise transition matrix is, and explain why the noise transition matrix is important.

\begin{definition}
(Noise transition matrix~\cite{van2017theory}) Suppose that the observed noisy label $\bar{y}$ is drawn independently from a corrupted distribution $p(X,\bar{Y})$, where features are intact. Meanwhile, there exists a corruption process, transition from the latent clean label $y$ to the observed noisy label $\bar{y}$. Such a corruption process can be approximately modeled via a \textit{noise transition matrix} $T$, where $T_{ij} = p(\bar{y} = e_j| y = e_i)$. 
\end{definition}

\begin{figure}[ht]
\centering
\subfigure[Sym-flipping.]
{\includegraphics[height=0.08\textheight]{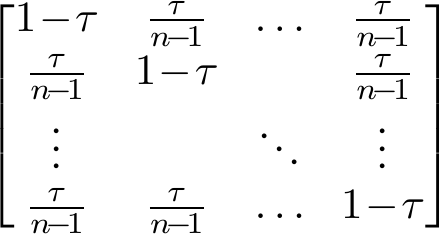}}
\qquad
\subfigure[Pair-flipping.]
{\includegraphics[height=0.084\textheight]{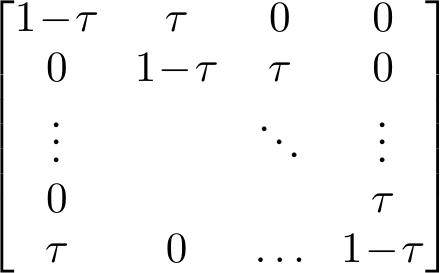}}
\vspace{-8px}
\caption{Two representatives of transition matrix $T$.}
\label{fig:repT}
\end{figure}

To further understand $T$, 
we present two representative 
structures of $T$ (Figure~\ref{fig:repT}): 
(1) Sym-flipping~\cite{van2015learning}; 
(2) Pair-flipping~\cite{han2018masking}. 
The definition of the corresponding $T$ is as follow,
where $\tau$ is the noise rate and $n$ is the number of the classes.

Specifically, the Sym-flipping structure models the common classification scenario under label noise, where the class of clean label can uniformly flip into other classes. Meanwhile, the Pair-flipping structure models the fine-grained classification scenario, where the class (e.g., Norwich terrier) of a clean label can flip into its adjunct class (e.g., Norfolk terrier) instead of a far-away class (e.g., Australian terrier). In the area of label-noise learning, we normally leverage the above two structures to generate simulated noise, and explore the root cause why the proposed algorithms can work on the simulated noise. Nonetheless, the real-world scenarios are very complex, where the noise transition matrix may not have structural rules (i.e., irregular). For example, \textit{Clothing1M}~\cite{xiao2015learning} is a Taobao clothing dataset, where mislabeled clothing images often share similar visual patterns. The noise structure of \textit{Clothing1M} is irregularly asymmetric, which is hard ti estimate.

The mathematical modeling of $T$ can be found in Section~\ref{sec:thm:idata} (Eq.~\eqref{eq:trans}),
which has been widely studied to build statistically consistent classifiers.
Normally, the clean class posterior can be inferred by using $T$ and the noisy class posterior, i.e., we have the important equation $p(\bar{Y}|x)=T \cdot p(Y|x)$, where  $T$ is a bridge between clean and noisy information.
As the noisy class posterior can be estimated by exploiting the noisy training data, 
the key step remains how to effectively estimate $T$ and leverage the estimated matrix to combat  label noise. 
Based on this observation, there are three general ways as 
in Sections~\ref{sec:data:ada}-\ref{sec:data:pro}.

\subsection{Adaptation Layer}
\label{sec:data:ada}

\begin{figure}[ht]
	\centering
	\includegraphics[width=0.35\textwidth]{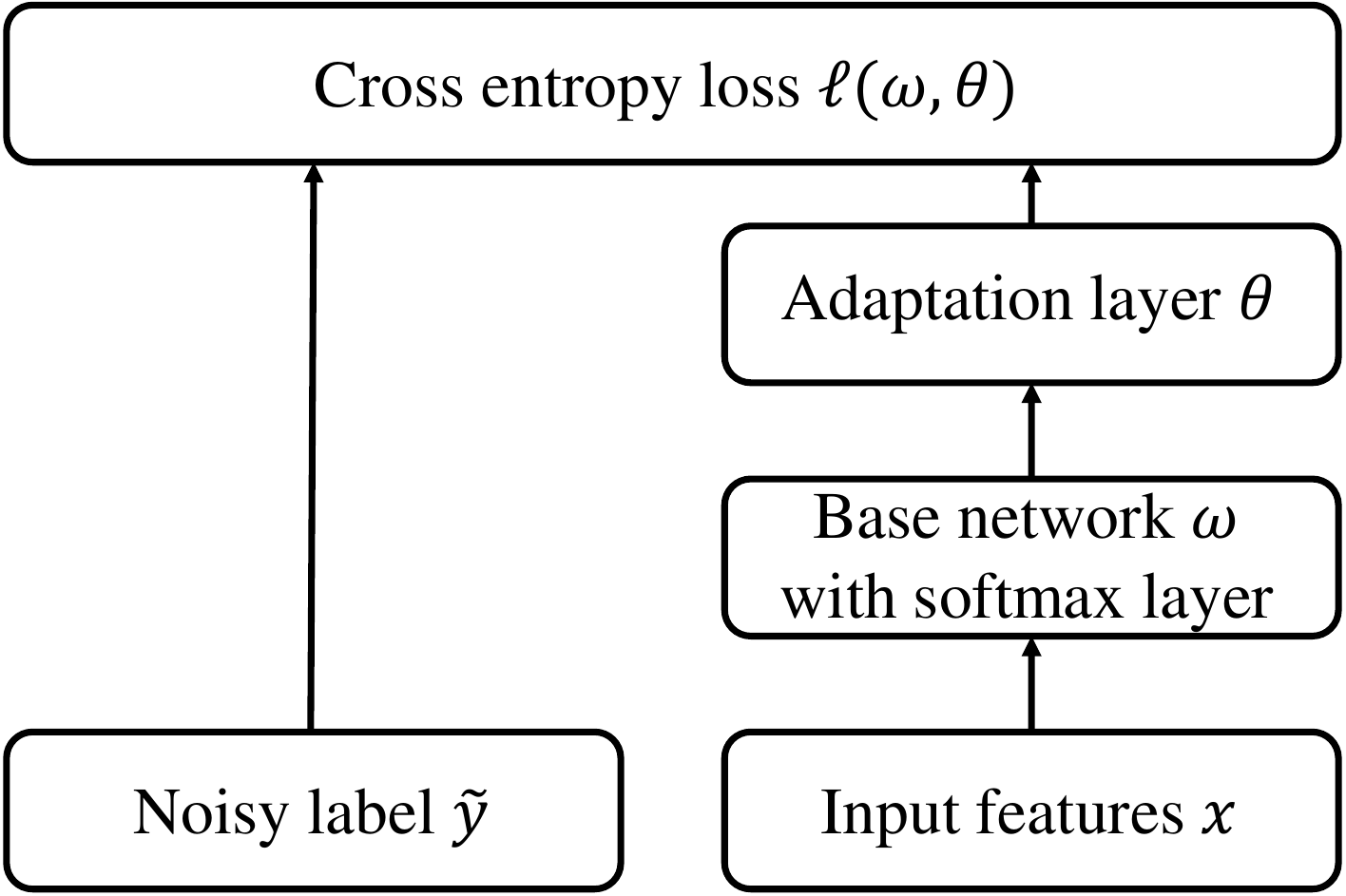}
	\vspace{-5px}
	\caption{A general case of adaptation layer.}
	\label{adpt-layer}
\end{figure}

Deep learning can be viewed as an end-to-end learning system. 
Therefore, the most intuitive way is to add an adaptation layer (Fig.~\ref{adpt-layer}) to estimate the transition matrix.

\subsubsection{Linear Case}
To realize this adaptation layer, Sukhbaatar et al.~\cite{sukhbaatar2014training} proposed a constrained linear layer (i.e., constrained to be a probability matrix) inserted between the base network and cross-entropy loss layer. This linear adaptation layer is parameterized by $T$, which is equivalent to the function of  $T$. 
Based on this idea, we can modify a classification model using a probability matrix $T$ that modifies its prediction to match the label distribution of the noisy data. 

The training model consists of two independent parts: the base model parameterized by $\omega$ and the noise model parameterized by $T$. Since the noise matrix $T$ has been modeled as a constrained linear layer,  update of the $T$ matrix can be easily carried out by back-propagating the cross-entropy loss.
However, it is hard to achieve the optimal $T$ via minimizing the cross-entropy loss, which is jointly parameterized by $\omega$ and $T$. 
To achieve the optimal $T$, Sukhbaatar et al.~\cite{sukhbaatar2014training} leveraged a regularizer on $T$, 
e.g., the trace norm or ridge regression, which forces it to approximate the optimal $T$. 
This work paves the way for deep learning with noisy labels, which directly motivates the following nonlinear case of the adaptation layer.

\subsubsection{Nonlinear Case}
Following the linear case, Goldberger et al.~\cite{goldberger2016training} proposed a non-linear layer inserted between the base network and cross-entropy loss layer to realize this adaptation layer.
Beyond the linear case, the training model consists of two independent parts: the base model parameterized by $\omega$ and the noise model/channel parameterized by $\theta$ (equal to the function of the noise transition matrix). Since the latent outputs 
of the base model are hidden, they proposed to leverage the expectation-maximization (EM) algorithm~\cite{dempster1977maximum} to estimate the hidden outputs (E-step) and the current parameters (M-step). Different from the linear case, the nonlinear case is free of strong assumptions (see Section 3.2 in~\cite{sukhbaatar2014training}).

However, there are several potential drawbacks to the EM-based approach, such as local optimality and scalability. To address these issues, Goldberger et al.~\cite{goldberger2016training} proposed two noise modeling variants: the c-model and s-model. Specifically, the c-model predicts the noisy label based on both the latent true label and the input features; while the s-model predicts the noisy label only based on the latent true label. Since the EM algorithm is equivalent to the s-model, they regarded both $\omega$ and $\theta$ as components of the same network and optimized them simultaneously. Moreover, the s-model is similar to the linear case proposed by Sukhbaatar et al.~\cite{sukhbaatar2014training}, although they proposed a different learning procedure. Note that though the c-model depends on the input features, and they still leverage network training in the M-step to update $\theta$.

\subsection{Loss Correction}
Another important branch is to conduct loss correction via leveraging  $T$, 
which can  also be integrated into deep learning systems. 
The aim of loss correction is that, training on noisy labels via the corrected loss should be approximately equal to training on clean labels via the original loss.
Note that we also introduce a label smoothing method related to loss correction in Section~\ref{label-smoothing}.

\subsubsection{Backward/Forward Correction}
\label{sec:bfcorr}

Patrini et al.~\cite{patrini2017making} introduced two loss correction techniques, namely forward correction and backward correction. 
The backward procedure corrects the loss by inverse transition matrix $T^{-1}$; 
while the forward procedure corrects the network predictions by $T$. 
Both corrections share the formal theoretical guarantees \textit{w.r.t} the clean data distribution.

\begin{theorem}
\label{bc-theorem} 
(Backward Correction, Theorem 1 in~\cite{patrini2017making}) Suppose that $T$ is non-singular, where $T_{ij} = p(\bar{y} = j| y = i)$ given that corrupted label $\bar{y} = j$ is flipped from clean label $y = i$. Given loss $\ell$ and network function $f$, Backward Correction is defined as
\begin{equation}\label{lemma-1-1}
  \ell^{\leftarrow}(f(x),\bar{y}) = [T^{-1}\ell_{y|f(x)}]_{\bar{y}},
\end{equation}
where $\ell_{y|f(x)} = (\ell(f(x),1),\ldots,\ell(f(x),k))$. Then, corrected loss $\ell^{\leftarrow}(f(x),\bar{y})$ is unbiased, namely,
\begin{equation}\label{lemma-1-2}
  \mathbb{E}_{\bar{y}|x}\ell^{\leftarrow}(f(x),\bar{y}) = \mathbb{E}_{y|x}\ell(f(x), y), \forall x.
\end{equation}
\end{theorem}

\begin{remark}
Backward Correction is unbiased. The LHS of (\ref{lemma-1-2}) is computed from the corrupted labels, and the RHS of (\ref{lemma-1-2}) is computed from the clean labels. Note that the corrected loss is differentiable, but not always non-negative~\cite{van2017theory}.
\end{remark}

\begin{theorem}\label{fw-theorem} (Forward Correction, Theorem 1 in~\cite{patrini2017making}) Suppose that the label
transition matrix $T$ is non-singular, where $T_{ij} = p(\bar{y} = j| y = i)$ given that corrupted label $\bar{y} = j$ is flipped from clean label $y = i$. Given loss $\ell$ and network function $f$, Forward Correction is defined as
\begin{equation}\label{lemma-2-1}
  \ell^{\rightarrow}(f(x), \bar{y}) = [\ell_{y|T^{\top}f(x)}]_{\bar{y}},
\end{equation}
where $\ell_{y|T^{\top}f(x)} = (\ell(T^{\top}f(x),1),\ldots,\ell(T^{\top}f(x),k))$. Then, the minimizer of the corrected loss under the noisy distribution is the same as the minimizer of the orginal loss under the clean distribution, namely,
\begin{equation}\label{lemma-2-2}
  \arg\min_f\mathbb{E}_{x,\bar{y}}\ell^{\rightarrow}(f(x), \bar{y}) = \arg\min_f\mathbb{E}_{x,y}\ell(f(x), y).
\end{equation}
\end{theorem}

\begin{remark}
For Forward Correction, the LHS of (\ref{lemma-2-2}) is computed from the corrupted labels, and the RHS of (\ref{lemma-2-2}) is computed from the clean labels. Note that the property is weaker 
than unbiasedness of Theorem~\ref{bc-theorem}, since the property is a necessary but not a sufficient condition of unbiasedness.
\end{remark}

Normally, $T$ is unknown, which needs to be estimated (i.e., $\widehat{T}$). Therefore, Patrini et al.~\cite{patrini2017making} proposed a robust two-stage training. First, they trained the network with $\ell$ on noisy data, and obtained an estimate of $T$ via the softmax output. Then, they re-trained the network with the corrected loss by $\widehat{T}$. Note that the estimation quality of $T$ directly decides the learning performance via the loss correction.

\subsubsection{Gold Correction}
Based on Forward Correction, 
Hendrycks et al.~\cite{hendrycks2018using} proposed Gold Loss Correction to handle severe noise. When severe noise exists, the transition matrix can not be estimated accurately by purely noisy data. 
The key motivation is to assume that a small subset of the training data is trusted and available.
Normally, a large number of crowdsourced workers may produce an untrusted set $\widetilde{D}$; 
while a small number of experts can produce a trusted set $D$. In a high-level, Hendrycks et al.~\cite{hendrycks2018using} aimed to leverage $D$ to estimate the $T$ accurately, 
and employed Forward Correction based on the estimated matrix. Then, they trained deep neural networks on $\widetilde{D}$ via the corrected loss, 
while training on $D$ via the original loss. 
This method is called Gold Loss Correction (GLC).

Thus, 
GLC's key step is to estimate $T$ accurately via a trusted set $D$. 
Specifically, 
we can estimate $T$ by $\widehat{T}$ as follows.
\begin{equation*}
\widehat{T}_{ij} 
= \frac{1}{A_i}\sum\nolimits_{x \in A_i}\widehat{p}(\bar{Y} = e_j|Y = e_i,x),
\end{equation*}
where $A_i$ is the subset of $x$ in $D$ with label $i$ and a classifier $\widehat{p}(\bar{y}|x)$ can be modeled by deep neural networks trained on $\widetilde{D}$. Empirically, the better estimate $\widehat{T}$ will lead to the better GLC's performance.

\subsubsection{Label Smoothing}\label{label-smoothing}
The technique of label smoothing is to smooth labels by mixing in a uniform label vector, whose label distribution belongs to the uniform distribution~\cite{szegedy2016rethinking}, which is a means of regularization. Lukasik et al.~\cite{lukasik2020does} relates label smoothing to a general family of loss-correction techniques, 
which demonstrates that label smoothing significantly improves performance under label noise. 
In general, \cite{szegedy2016rethinking} and \cite{lukasik2020does} can be unified into a label smearing framework:
\begin{equation*}
\ell^{\text{SM}}(f_{\theta}(X),Y) 
= M \cdot \ell(f_{\theta}(X),Y),
\end{equation*}
where $M$ is a smearing matrix~\cite{frodesen1979probability}. Such a matrix is used for bridging the original loss $\ell$ and the smeared loss $\ell^{\text{SM}}$.
Therefore, in this framework, there are three examples:
\begin{itemize}[leftmargin=*]
\item Standard training: suppose $M = I$, where $I$ is the identity matrix.
\item Label smoothing: suppose $M = (1-\alpha) I + \frac{\alpha J}{L} $, where $J$ is the all-ones matrix and $L$ is the number of classes.
\item Backward correction under symmetric noise: suppose $M = \frac{1}{1-\alpha}\cdot(I - \frac{\alpha J}{L})$, where $M = T^{-1}$ in Theorem~\ref{bc-theorem}. 
\end{itemize}
We can clearly see the close connection between label smoothing and backward correction. Actually, label smoothing can have a similar effect to 
shrinkage regularization~\cite{wasserman2013all}.

\subsection{Prior Knowledge}
\label{sec:data:pro}

As mentioned before, 
methods in this section solve the LNRL problem by estimating $T$. 
However, its accurate estimation can be hard in real-world scenarios, 
which motivates 
researchers
to incorporate prior knowledge for better estimation.

\subsubsection{Human-in-the-Loop Estimation}
Han et al.~\cite{han2018masking} proposed a human-assisted approach called ``Masking'', which decouples the structure and value of the transition matrix. Specifically, the structure can be viewed as a prior knowledge, coming from human cognition, since human can mask invalid class transitions (e.g., cat $\nleftrightarrow$ car). Given the structure information, we can only focus on estimating the noise transition probability along the structure in an end-to-end system. 
Therefore, the estimation burden will be largely reduced. Actually, it makes sense that human cognition masks invalid class transitions and highlights valid class transitions, such as the column-diagonal, tri-diagonal and block-diagonal structures. Therefore, the remaining issue is how to incorporate such prior structure into an end-to-end learning system. The answer is a generative model.

Specifically, there are three steps in Masking. 
First, the latent ground-truth label is from $y\sim p(y|x)$, 
where $p(y|x)$ is a categorical distribution. Second, the noise transition matrix variable $t$ is from $t\sim p(t)$  
and its structure 
is generated as $t_o\sim p(t_o)$, 
where $p(t)$ is an implicit distribution modeled by a neural network without its explicit form, structure prior $p(t_o)=p(t)\frac{dt}{dt_o}\big|_{t_o=f(t)}$ provided by human cognition, 
and $f(\cdot)$ is the mapping function from $t$ to $t_o$. 
Third, the noisy label is from $\bar{y}\sim p(\bar{y}|y,t)$, where $p(\bar{y}|y,t)$ models the transition from $y$ to $\bar{y}$ given $t$. Based on this generative process, 
we can deduce the evidence lower bound~\cite{wainwright2008graphical}
to approximate the log-likelihood of the noisy data.

\subsubsection{Fine-tuning Revision}
Xia et al.~\cite{xia2019anchor} introduced a transition-revision method to effectively learn the transition matrix, 
which is called Reweight T-Revision (Reweight-R). 
Specifically, they first initialized the transition matrix by exploiting data points that are similar to anchor points, having high noisy class posterior probabilities. Thus, in the Reweighted-R method, the initialized transition matrix is viewed as a prior knowledge. Then, they fine-tuned the initialized matrix by adding a slack variable, which is then learned and validated together with the classifier by using noisy data.

Specifically, given noisy training sample 
$\bar{\mathcal{D}}^{\text{tr}}$ and noisy validation set $\bar{\mathcal{D}}^{\text{v}}$, 
there are two stages in Reweight-R. In the first stage, the unweighted loss is minimized to learn $\hat{p}(\bar{Y}|X=x)$ without a noise adaption layer. 
Then, the noise transition matrix $\hat{T}$ is initialized, 
which can be viewed as a prior knowledge for further fine-tuning. 
Namely, $\hat{T}$ is initialized according to (1) in \cite{xia2019anchor} by using data with the highest $\hat{p}(\bar{Y} = e_i|X=x)$ as anchor points for the $i$-th class.
In the second stage, based on the prior $\hat{T}$, the neural network is initialized by minimizing the weighted loss with a noisy adaptation layer $\hat{T}^\top$. 
Furthermore, the weighted loss is minimized to learn classifier $f$ and incremental $\Delta T$ with a noisy adaptation layer $(\hat{T} + \Delta T)^\top$. 
Namely, the second stage modifies $\hat{T}$ gradually by adding a slack variable $\Delta T$, 
and learns the classifier and $\Delta T$ by minimizing the weighted loss. 
The two stages alternate until converges, namely achieving minimum error on $\bar{\mathcal{D}}^{\text{v}}$.

\section{Objective}
\label{sec:obj}

Methods in this section solve the LNRL problem by modifying the objective function $\tilde{\ell}$ in \eqref{eq:noisy-risk}, 
and such modification can be realized in 
three different ways. 
The first way is to directly augment the objective function by either explicit regularization or implicit regularization. 
Note that  implicit regularization tends to operate at the algorithm level, 
which is equivalent to modifying the objective function. 
The second way is to assign dynamic weights to different objective sub-functions, 
and a larger weight means higher importance for the corresponding sub-function. The last way is to directly 
redesign new loss functions.

\subsection{Regularization}

Regularization is the most direct way to modify the objective function. 
Mathematically, we  add a regularization term to the original objective, i.e., $\tilde{\ell} = \ell + r$. 
In label-noise learning, the aim of regularization is to achieve better generalization, which avoids or alleviates overfitting noisy labels.
There are two intuitive ways as follows.

\subsubsection{Explicit Regularization}
\label{exp-reg}
Azadi et al.~\cite{azadi2015auxiliary} proposed a novel regularizer $r = \Omega_{\text{aux}}(w)$ to exploit the data structure for combating label noise, 
where $\Omega_{\text{aux}}(w) = \left\lVert Fw\right\rVert_\mathrm{g}$. 
Note that $\left\lVert \cdot \right\rVert_\mathrm{g}$ denotes the group norm 
and $F^{\top} = [X_1,\ldots,X_n]$ represents the set of diagonal matrices (e.g., $X_i$) consisting of the (e.g., $i$-th) data features,
which induces a non-overlapping group sparsity
that encourages  most coefficients to be zero. 
This operation will encourage a small number of clean data to contribute to learning of the model, while filtering mislabeled and non-relevant data. 
In other words, this regularizer enforces the features of the good data to be used for modeling learning, while noisy additional activations will be disregarded. 

It is worth noted that Berthelot et al.~\cite{berthelot2019mixmatch} introduced MixMatch for semi-supervised learning (SSL), and the empirical performance of MixMatch reaches the state-of-the-art. Most importantly, one of the key components in MixMatch is Minimum Entropy Regularization (MER), which belongs to explicit regularization in LNRL. Speicifically, MER was proposed by Grandvalet \& Bengio~\cite{grandvalet2005semi}, and the key idea is to augment the cross-entropy loss with an explicit term encouraging the classifier to make predictions with high confidence on the unlabeled examples, namely minimizing the model entropy  
for unlabeled data $x$. Similar to MER, the pseudo-label method conducts entropy minimization implicitly~\cite{lee2013pseudo}, which generates hard labels from high-confidence predictions on unlabeled data for further training. In particular, the pseudo-label method (i.e., label guessing) first computes the average of the model’s predicted class distributions across all augmentations, and then applies a temperature sharpening function to reduce the entropy of the label distribution.

Miyato et al.~\cite{miyato2018virtual} also explored the smoothness for combating label noise, and they proposed the virtual adversarial loss, which is a new measure of local smoothness of the conditional label distribution given input. Specifically, their method enforces the output distribution to be isotropically smooth around each input data 
via selectively smoothing the model in its most anisotropic direction. The
benefit of such smoothness makes the model output insensitive  to the input data. To realize such smoothness, they first design the virtual adversarial direction, which can greatly deviate from the current inferred output distribution, from the status quo without the label information. Based on such a direction, they defined local distributional smoothness, and then proposed a training method called virtual adversarial training (VAT).
Mathematically, 
they first defined local distributional smoothness (LDS):
\begin{equation}
\text{LDS}(x^*,\theta):=D[p(y|x^*,\hat{\theta}),p(y|x^*+r_{\text{vadv}},\theta)],
\end{equation}
where
$r_{\text{vadv}}
:= \arg\max\nolimits_{\left\lVert r \right\rVert_2 \leq \epsilon} 
D[ p(y|x^*,\hat{\theta}),p(y|x^*+r) ]$,
$D$ is a non-negative function that measures the distribution divergence, and
$x^*$ represents either labeled or unlabeled features. We use $\hat{\theta}$ to
denote the vector of model parameters at a specific iteration step of the
training process. Then, we have a regularization term:
\begin{equation*}
R_{\text{vadv}}(D_\mathrm{l},D_\mathrm{ul},\theta)
:=\nicefrac{1}{N_l+N_{ul}}
\sum\nolimits_{x^* \in D_l,D_{ul}}\text{LDS}(x^*,\theta),
\end{equation*}
where $D_l$ denotes the label dataset with size $N_l$; and $D_{ul}$ denotes the unlabeled dataset with size $N_{ul}$. 
Therefore, the full objective function of VAT is given by
\begin{equation*}
\ell(D_l,\theta) + \alpha R_{\text{vadv}}(D_l,D_{ul},\theta).
\end{equation*}

\subsubsection{Implicit Regularization}
Recently, there are more and more implicit regularizers, which take the effects of regularization without the explicit form (cf.~Section~\ref{exp-reg}).
For example,

\begin{itemize}[leftmargin=*]
\item Bootstrapping~\cite{reed2014training}: Reed et al.~\cite{reed2014training}
augmented the prediction objective with a notion of consistency. In the
high-level, this provides the learner justification to ``disagree'' with a
perceptually-inconsistent training label, and effectively re-label the data.
Namely, the learner bootstraps itself in this way, which uses a convex
combination of training labels and the current model’s predictions to generate
the training targets. Intuitively, as the learner improves over time, its
predictions can be trusted more. Such bootstrapping can avoid modeling the noise
distribution. Specifically, Reed et al.~\cite{reed2014training} proposed two ways
to realize bootstrapping, such as soft and hard bootstrapping. The soft version
loss $\ell_{\text{soft}}$ uses predicted class probabilities $q$ directly to
generate regression targets for each batch as follows:
\begin{equation*}
\ell_{\text{soft}}(q,t) 
= \sum\nolimits_{k=1}^{L}[\beta t_k + (1-\beta)q_k]\log(q_k),
\end{equation*}
where $t$ denotes the training labels.
The parameter $\beta$ balances the prediction $q$ and target $t$, which can be found via cross validation. The soft version loss is equivalent to softmax regression with minimum entropy regularization, which encourages the model to have a high confidence in predicting labels. Meanwhile, the hard version loss $\ell_{\text{hard}}$ modifies targets using a MAP estimate of $q$ given $\mathbf{x}$ as follows:
\begin{equation*}
\ell_{\text{hard}}(q,t) 
= \sum\nolimits_{k=1}^{L}[\beta t_k + (1-\beta)z_k]\log(q_k),
\end{equation*}
where $z_k = \mathbf{1}[k=\arg\max_{i = 1,\ldots,L} q_i]$ and $\mathbf{1}[\cdot]$ denotes the indicator function. To solve the hard version via stochastic gradient descent, an EM-like method will be employed. In the E-step, the approximate-truth confidence targets are estimated as a convex combination of training labels and model predictions. In the M-step, the model parameters are updated in order to predict those generated targets better.

\item Mixup~\cite{zhang2017mixup}: Motivated by vicinal risk minimization~\cite{chapelle2000vicinal}, Zhang et al.~\cite{zhang2017mixup} introduced a data-agnostic regularization method called Mixup, which constructs virtual training examples $(\tilde{x},\tilde{y})$ as follows.
\begin{equation}\label{mixup}
\tilde{x} = \lambda x_i + (1-\lambda)x_j
\;\text{and}\;
\tilde{y} = \lambda y_i + (1-\lambda)y_j,
\end{equation}
where $(x_i,y_i)$ and $(x_j,y_j)$ are two examples randomly drawn from the training data and $\lambda \in [0,1]$ is a balanced parameter between two examples.
Intuitively, Mixup conducts virtual data augmentation, 
which dilutes the noise effects and smooths the data manifold. 
This simple but effective idea can be used in not only noisy labels but also adversarial examples, 
since the smoothness happens in both features and labels according to \eqref{mixup}.

\item SIGUA~\cite{han2020sigua}: It is noted that, given data with noisy labels, 
over-parameterized deep networks can gradually fit the whole data~\cite{zhang2016understanding,arpit2017closer,yao2020s2e}.
To relieve this issue, 
Han et al.~\cite{han2020sigua} proposed a gradient-ascent method called stochastic integrated gradient underweighted ascent (SIGUA): 
in a mini-batch, 
we adopt gradient descent on good data as usual, and learning-rate-reduced gradient ascent on bad data; 
the proposal is a versatile approach where data goodness or badness is w.r.t. desired or undesired memorization given a base learning method.
Technically, SIGUA is a specially designed regularization by pulling optimization back for generalization when their goals conflict with each other.
A key difference between SIGUA and parameter shrinkage like weight decay is that SIGUA pulls optimization back on some data but parameter shrinkage does the same on all data; philosophically, SIGUA shows that forgetting undesired memorization can reinforce desired one, which provides a novel viewpoint on the inductive bias of neural networks.
\end{itemize}

\subsection{Reweighting}
Instead of augmenting the objective via explicit/implicit regularization, 
there is another typical solution to modify the objective function termed Reweighting. 
In high-level,
Reweighting is a way to assign different weights to different sub-objective 
functions, where each sub-objective function corresponds to each training sample.
The weights should be learned, and the larger weights will bias sub-objectives to better overcome label noise. 
The procedure can be formulated as $\tilde{\ell} = \sum_i w_i\ell_i$. Here, we introduce several Reweighting approaches as follows.

\subsubsection{Importance Reweighting}
Liu and Tao \cite{liu2015classification} introduced importance reweighting \cite{gretton2009covariate} from domain adaptation to label-noise learning by treating the noisy training data as the source domain and the clean test data as the target domain. The idea is to rewrite the risk w.r.t. the clean data by exploiting the noisy data. Specifically, 
\begin{align*}
\label{eq:importance}%
&R(f)=\mathbb{E}_{(X,Y)\sim D}[\ell(f(X),Y)]\\
&=\int_{x}\sum_ip_D(X=x,Y=i)\ell(f(x),i)dx\nonumber\\
&=\int_{x}\sum_ip_{\bar{D}}(X=x,\bar{Y}=i)\frac{p_D(X=x,\bar{Y}=i)}{p_{\bar{D}}(X=x,\bar{Y}=i)}\ell(f(x),i)dx\nonumber\\
&=\int_{x}\sum_ip_{\bar{D}}(X=x,\bar{Y}=i)\frac{p_D(\bar{Y}=i|X=x)}{p_{\bar{D}}(\bar{Y}=i|X=x)}\ell(f(x),i)dx\\
&=\mathbb{E}_{(X,\bar{Y})\sim \bar{D}}[\beta(X,\bar{Y}){\ell}(f(X),\bar{Y})],\nonumber
\end{align*}
where the second last equation holds because label noise is assumed to be independent of instances and $\beta(X,\bar{Y})=p_D(\bar{Y}=i|X=x)/p_{\bar{D}}(\bar{Y}=i|X=x)$ denotes the weights which plays a core role in importance reweighting and can be learned by either exploiting the transition matrix or a small set of clean data.

\subsubsection{Bayesian Method}
Wang et al.~\cite{wang2017robust} proposed reweighted probabilistic models (RPM) to combat label noise. 
The idea is simple and intuitive: down-weighting on corrupted labels but up-weighting on clean labels, which brings us Bayesian data reweighting. The mathematical formulations include three steps:
\begin{itemize}[leftmargin=15px]
\item[1)] Define a probabilistic model $p_{\beta}(\beta)\prod_{n=1}^{N}\ell(y_n|\beta)$.

\item[2)] Assign a positive latent weight $w_n$ to each likelihood $\ell(\cdot|\beta)$,
and choose a prior on the weights $p_w(w)$, where $w=(w_1,\ldots, w_N)$. Thus, the RPM can be represented by
\begin{equation*}
p(y,\beta,w)
= \nicefrac{1}{Z} \cdot p_{\beta}(\beta)p_{w}(w)
\prod\nolimits_{n=1}^{N}\ell(y_n|\beta)^{w_n},
\end{equation*}
where $Z$ is the normalizing factor.

\item[3)] Infer the posterior of both the latent variables $\beta$ and the weight $w$: $p(\beta,w|y)$. The prior knowledge on the weights $p_w(w)$ trades off extremely low likelihood terms, where the options of $p_w(w)$ are a bank of Beta priors, a scaled Dirichlet prior and a bank of Gamma priors. Note that RPMs treat weights $w$ as latent variables, which are automatically inferred.
\end{itemize}


Arazo et al.~\cite{arazo2019unsupervised} introduced a two-component (clean-noisy) beta mixture model (BMM) for a mixture of clean and noisy data, which brings us a bootstrapping loss. Specifically, the posterior probabilities under BMM are leveraged to implement a dynamically-weighted bootstrapping loss, robustly dealing with noisy samples without discarding them. Mathematically, the probability density function 
of a mixture model of $K$ components 
on the loss $\ell$ is defined as
\begin{equation}
p(\ell)
= \sum\nolimits_{k=1}^K \lambda_k p(\ell|k),
\end{equation}
where $K=2$ is our case, $\lambda_k$ are mixing 
weights, and $p(\ell|k)$ can be modeled by the beta distribution:
\begin{equation}
p(\ell|\alpha,\beta)
= \frac{\Gamma(\alpha+\beta)}{\Gamma(\alpha)\Gamma(\beta)} \ell^{\alpha-1} (1-\ell)^{\beta-1}.
\end{equation}
The above BMM can be solved by the EM procedure. Specifically, latent variables $\gamma_k(\ell) = p(k|\ell)$ are introduced. In the E-step,  $\lambda_k$, $\alpha_k$, $\beta_k$ are fixed and $\gamma_k(\ell)$ is updated via the Bayes rule. In the M-step, given fixed $\gamma_k(\ell)$, $\alpha_k$ and $\beta_k$ are estimated using the weighted moments. Meanwhile, the dynamic weights are updated in an easy way: $\lambda_k = \frac{1}{N}\sum_{i=1}^{N}\gamma_k(\ell_i)$. Based on this BMM model, they further proposed dynamic hard/soft bootstrapping losses, where the weight $w_i$ of each sample is dynamically set to $p(k=1|\ell_i)$.

\subsubsection{Neural Networks}
Shu et al.~\cite{shu2019meta} introduced Meta-Weight-Net (MW-Net), which can adaptively learn an explicit weighting function from data. 
In high-level, 
the weighting function is an MLP with one hidden layer, mapping from a loss to weights. 
Mathematically, the optimal parameter $w$ can be calculated by minimizing the weighted loss.
\begin{align*}
w^*(\theta) 
= \arg\min_{w}\mathcal{\ell}^{\text{tr}}(w;\theta)
= \nicefrac{1}{N}\sum\nolimits_{i=1}^{N}\mathcal{V}(\ell_i^{\text{tr}}(w);\theta)\ell_i^{\text{tr}}(w),
\end{align*}
where $\mathcal{V}(\ell_i^{\text{tr}}(w);\theta)$ denotes MW-Net.
Here, the parameters in MW-Net can be optimized by the meta learning idea. Given a small amount of clean and balanced meta-data $\{x_i^{(\text{meta})},y_i^{(\text{meta})}\}_{i=1}^M$, the optimal parameter $\theta$ can be obtained by minimizing the meta-loss:
\begin{align*}
\theta^* 
= \arg\min_{\theta}\ell^{\text{meta}}(w^*(\theta))
=\nicefrac{1}{M}\sum\nolimits_{i=1}^{M}\ell_i^{\text{meta}}(w^*(\theta)).
\end{align*}
Then SGD is employed to update $w$ (parameters of classifier network) and $\theta$ (parameters of MW-Net) iteratively.

\subsection{Redesigning}
Besides augmenting and reweighting the objective function, there is another common solution called redesigning. 
In high-level, redesigning,
which generally replaces $\tilde{\ell}$ with a special format $\ell'$
different  from $\ell$,
is motivated by different observations and principles.
Thus,
these methods are diverse much for different scenarios. Here, we introduce several redesigning approaches as follows.

\subsubsection{Loss Redesign}
In recent years, there are a lot of 
new losses for combating label noise. Their designs are based on different principles, such as gradient clipping~\cite{menon2019can} and curriculum learning~\cite{bengio2009curriculum}. 
Here, we choose several representative losses to explain.
\begin{itemize}[leftmargin=*]
\item Zhang et al.~\cite{zhang2018generalized} proposed a generalized cross-entropy loss called $\ell_q$, which encompasses both the mean absolute error
(MAE) and categorical cross entropy (CCE) loss. Since the $\ell_q$ loss is a generalization of MAE and CCE, it enjoys the benefits of both the noise-robustness provided by MAE and the implicit weighting scheme of CCE, which can be well justified by theoretical analysis.
More importantly, it empirically works well for both closed-set noise~\cite{han2018co} and open-set noise~\cite{wang2018iterative}.
Mathematically, they used the negative Box-Cox transformation as a 
$\ell_q$ loss function: 
\begin{equation*}
\ell_q(f(x),e_j) = (1-f_j(x)^q) / q,
\end{equation*}
where $q \in (0,1]$ and $e_j$ is one-hot vector belonging to ghe $j$-th class. The $\ell_q$ loss is reduced to CCE when
$\lim_{q\rightarrow0}\ell_q(f(x),e_j)$, and becomes MAE when $q = 1$. Since a
tighter bound in $\ell_q$ can bring stronger noise tolerance due to discarding many samples for training, they proposed the truncated $\ell_q$ loss:
\begin{equation}
\ell_\text{trunc}(f(x),e_j)
=
\begin{cases}
\ell_q(k) & \text{if} \; f_j(x) \leq k, \\
\ell_q(f(x),e_j) & \text{otherwise}, \\
\end{cases}
\end{equation}
where $0 < k < 1$, and $\ell_q(k) = \nicefrac{1-k^q}{q}$. When $k \rightarrow 0$, the truncated $\ell_q$ loss equals the original $\ell_q$ loss.

\item
Charoenphakdee et al.~\cite{charoenphakdee2019symmetric} theoretically justified symmetric losses through the lens of theoretical tools, including the classification-calibration condition, excess risk bound, conditional risk minimizer and AUC-consistency. The key idea is to design a loss that does not have to satisfy the symmetric condition everywhere, which gives a high penalty in the non-symmetric area,
i.e., $\ell(z) + \ell(-z)$ is a constant for every $z \in \mathbb{R}$. Motivated by this phenomenon, they introduced a barrier hinge loss, satisfying a symmetric condition not everywhere but gives a large penalty once $z$ is outside of the symmetric interval,
which incentivizes to learn a prediction function inside of the symmetric interval. 
Mathematically, a barrier hinge loss is defined as follows.
\begin{equation}
\ell(z) = \max(-b(r+z)+r,\max(b(z-r),r-z)),
\end{equation}
where $b > 1$ and $r > 0$.

\item
Thulasidasan et al.~\cite{thulasidasan2019combating} proposed to abstain some confusing examples during training deep networks. 
In practice, abstention has some relation with loss redesign. Based on abstention, they introduced a deep abstaining classifier (DAC). 
DAC has an additional output $p_{k+1}$, which indicates the probability of abstention.
The loss of DAC is as follows.
\begin{equation}
\ell(x_j) 
= - \tilde{p}_{k + 1} \sum\nolimits_{i=1}^k t_i\log( \nicefrac{p_i}{\tilde{p}_{k + 1}} )
- \alpha\log \tilde{p}_{k + 1},  
\end{equation}
where $\tilde{p}_{k + 1} = 1-p_{k+1}$. If $\alpha$ is large, the penalty drives $p_{k+1}$ to zero, which leads the model not to abstain. If $\alpha$ is small, the classifier may abstain on everything. They further proposed an auto-tuning algorithm to find the optimal $\alpha$. Note that DAC can be used for both structured (e.g., asymmetric and pair-flipping) and unstructured (i.e., symmetric) label noise, where DAC works as a data cleaner.

\item
Aditya et al.~\cite{menon2019can} leveraged gradient clipping to design a new loss. 
Intuitively, clipping the gradient prevents over-confident descent steps in the scenario of label noise. Motivated by gradient clipping, they proposed the partially Huberized loss
\begin{equation}
\tilde{\ell}_{\theta}(x,y)
\! = \!
\begin{cases}
-\tau p_{\theta}(x,y)
\! + \! \log\tau \! + \! 1 & \text{if} \; p_{\theta}(x,y) \leq \frac{1}{\tau}, 
\\
-\log p_{\theta}(x,y) & \text{otherwise}.
\end{cases}
\end{equation}

\item
Lyu et al.~\cite{lyu2019curriculum} proposed the curriculum loss (CL), which is a tighter 
upper bound of the 0-1 loss compared to the conventional surrogate of the 0-1 loss (cf. Eq. (8) in \cite{lyu2019curriculum}). Moreover, CL can adaptively select samples for stagewise training. In particular, giving any base loss function $\ell(u) \geq \mathbf{1} (u < 0), u \in \mathbb{R}$, CL is defined as follows.
\begin{align*}
Q(\mathbf{u}) 
=
\max
\left( 
\min\nolimits_{\mathbf{v}\in\{0,1\}^n} f_1(\mathbf{v})
,
\min\nolimits_{\mathbf{v}\in\{0,1\}^n} f_2(\mathbf{v})
\right),
\end{align*}
where
\begin{align*}
f_1(\mathbf{v}) 
\! = \! \sum_{i=1}^n v_i\ell(u_i)
\;\text{and}\;
f_2(\mathbf{v})
\! = \! n \! - \! \sum_{i=1}^n v_i \! + \! \sum_{i=1}^n \mathbf{1}(u_i \! < \! 0).
\end{align*}
To adapt CL for deep learning models, they further introduced the noise pruned CL.
\end{itemize}

\subsubsection{Label Ensemble}

\begin{itemize}[leftmargin=*]
\item Laine and Aila~\cite{laine2016temporal} introduced self-ensembling in semi-supervised learning, including the $\pi$-model and temporal ensembling, which can also be used to purify the label noise. The key idea of self-ensembling is to form a consensus prediction of the unknown labels using the outputs of the network in training. Specifically, the $\pi$-model encourages consistent network output between two realizations of the same input, under two different dropout conditions. 
Beyond the $\pi$-model, temporal ensembling is considered for the network predictions over multiple previous training epochs. 
The loss function of the $\pi$-model is
\begin{equation}
\ell = -\nicefrac{1}{B}\sum\nolimits_i
\log z_i[y_i] + \nicefrac{w(t)}{C|B|}
\sum\nolimits_i \left\|z_i-\tilde{z}_i\right\|^2,
\end{equation}
where the first term handles labeled data via the standard cross-entropy loss, namely, $\log z_i[y_i]$ calculates the cross-entropy loss value between model prediction $z_i$ and label $y_i$, and the second term handles unlabeled data. Note that $(x_i,y_i)$ denotes the pair of input and label, $B$ denotes the mini-batch size, and $C$ denotes the number of different classes. Both $z_i$ and $\tilde{z}_i$ are transformed from the same input $x_i$, namely two predictions via the same network with different dropout conditions. The second term is also scaled by time-dependent weighting function $w(t)$. Temporal ensembling goes beyond $\pi$-model by aggregating the predictions of multiple previous network evaluations into an ensemble prediction. Namely, the main difference from $\pi$-model is that the network and augmentations are evaluated only once per input in each epoch, and the target $\tilde{z}$ is based on prior network evaluations. 
After every training epoch, 
the network outputs $z_i$ are accumulated into ensemble outputs $Z_i$ by updating $Z_i \leftarrow \alpha Z_i + (1-\alpha)z_i$ and $\tilde{z} \leftarrow \nicefrac{Z_i}{(1-\alpha^t)}$, where $\alpha$ is a momentum term.

\item
Nguyen et al.~\cite{nguyen2019self} proposed a self-ensemble label filtering (SELF) method to progressively filter out the wrong labels during training. 
In high-level, they leveraged the knowledge provided in the network's output over different training iterations to form a consensus of predictions, which progressively identifies and filters out the noisy labels from the labeled data. In the filtering strategy, the model can determine the set of potentially corrected samples $L_i$ based on agreement between the label $y$ and its maximum likelihood prediction $\hat{y}_x$ with $L_i = \{(y,x)|\hat{y}_x = y; \forall (y,x) \in L_0\}$,
where $L_0$ is the sample set 
in the beginning. In the self-ensemble strategy, they maintained the two-level ensemble. First, they leveraged the model ensemble with Mean Teacher~\cite{tarvainen2017mean}, 
namely an exponential running average of model snapshots. Second, they employed a
prediction ensemble by collecting the sample predictions over multiple training
epochs: $\bar{z}_j = \alpha \bar{z}_{j-1} + (1-\alpha)\hat{z}_j$, where
$\bar{z}_j$ depicts the moving-average prediction of sample $k$ at epoch $j$ and
$\hat{z}_j$ is the model prediction for sample $k$ at epoch $j$.

\item
Ma et al.~\cite{ma2018dimensionality} 
investigated the dimensionality of the deep representation subspace of training samples. 
Then, they developed a dimensionality-driven learning strategy, 
which monitors the dimensionality of subspaces during training and adapts the loss function accordingly. The key idea is to leverage the local intrinsic dimensionality (LID), which discriminates clean labels and noisy labels during training deep networks.
Mathematically, the estimation of LID can be defined as
\begin{equation}
\widehat{\text{LID}} 
= -
\big( \nicefrac{1}{k}\sum\nolimits_{i=1}^k\log\nicefrac{r_i(x)}{r_{\max}(x)}\big)^{-1},
\end{equation}
where $r_i(x)$ denotes the distance between $x$ and its $i$-th nearest neighbor, and $r_{\max}(x)$ denotes the maximum of the neighbor distance. Specifically, when learning with clean labels, the LID score is consistently decreasing and the test accuracy is increasing with the increase of training epochs. However, when learning with noisy labels, the LID score first decreases and then increases after a few epochs. In contrast, the test accuracy is totally opposite, which first increases and then decreases. Based on the LID score, the dynamics of deep networks is overseen.

\end{itemize}

\section{Optimization Policy}
\label{sec:opt}

Methods in this section solve the LNRL problem by changing optimization policies, such as early stopping. The effectiveness of early stopping is due to memorization effects of deep neural networks, which avoid overfitting noisy labels to some degree. To combat with noisy labels using memorization effects, there exists another and possibly better way, namely small-loss tricks.
The structure of this section is arranged as follows. First, we explain what 
memorization effects are and why this phenomenon is important. Then, we introduce
several common ways to leverage  memorization effects for combating label
noise. The first common way is to self-train a single network robustly via
small-loss tricks, which brings us MentorNet~\cite{jiang2018mentornet} and Learning to
Reweight~\cite{ren2018learning}. Furthermore, the second common way is to co-train two
networks robustly via small-loss tricks, which brings us Co-teaching~\cite{han2018co} and Co-teaching+~\cite{yu2019does}. Lastly, there are several ways to
further improve the performance of Co-teaching, such as by using cross-validation~\cite{chen2019understanding},
automated learning~\cite{yao2020s2e} and   Gaussian mixture model~\cite{li2020dividemix}.

\subsection{Memorization Effects}
Arpit et al.~\cite{arpit2017closer} introduced a very critical work called ``A
closer look at memorization in deep networks'', which shapes a new direction
towards solving LNRL. In general, memorization effects can be defined as the
behavior exhibited by deep networks trained on noise data. Specifically, deep
networks tend to memorize and fit easy (clean) patterns, and gradually over-fit
hard (noisy) patterns~\cite{arpit2017closer}.  Here, we empirically reproduce a simulated experiment to justify this hypothesis, and experimental details can be found in Appendix 3.

\begin{figure}
\begin{center}
\centerline{\includegraphics[width=0.35\textwidth]{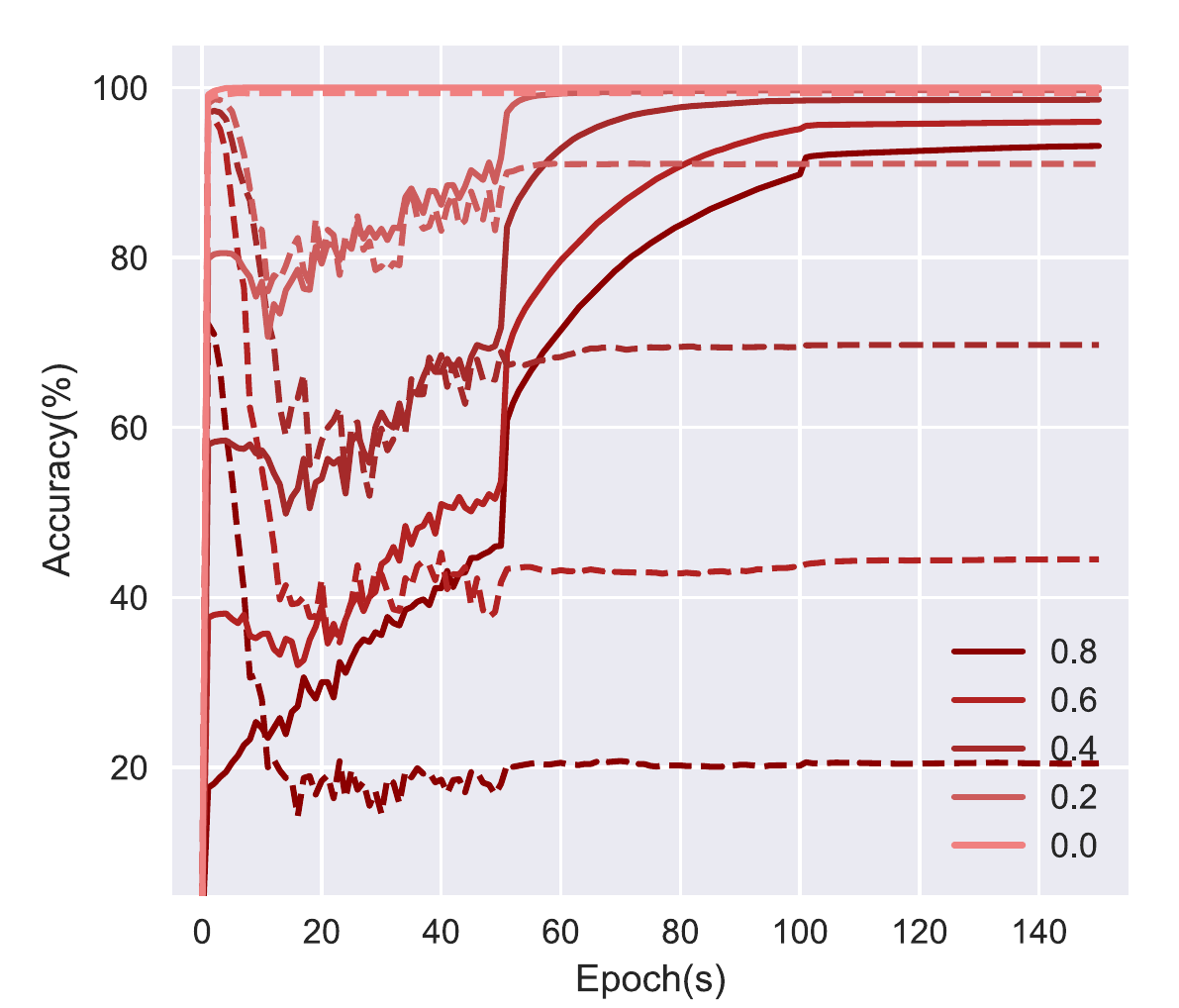}}
\vspace{-10px}
\caption{A simulated experiment based on different noise rates ($0\%$-$80\%$). We chose \textit{MNIST} with uniform noise as noisy data. The solid lines denote the training accuracy; while the dotted lines mean the validation accuracy.}
\label{sim-exp}
\end{center}
\end{figure}

In Fig.~\ref{sim-exp}, we used the MNIST dataset, and added random noise on its labels. 
The noise rate was chosen from the range between $0\%$ and $80\%$. We trained our deep networks on the corrupted training data. 
Then, we tested the trained networks on both (noisy) training data and
(clean) validation data. 
We can clearly see two phenomena in the graph:
\begin{itemize}[leftmargin=*]
    \item The training curve will  reach or approximate 100\% accuracy eventually. All curves will converge the same.
    \item The validation curve will first reach a very high accuracy in 
    the first few epochs, 
    but drop gradually until convergence (after 40 epochs).
\end{itemize}

Since deep networks tend to memorize and fit easy
(clean) patterns in the corrupted data, the validation curve will first reach a peak.
However, such overparameterized models will gradually over-fit hard (noisy)
patterns. The validation curve will drop gradually, since the validation data is
clean. This simple experiment not only justifies the hypothesis of memorization
effects, but also opens a new door to the LNRL problem, namely small-loss
tricks~\cite{jiang2018mentornet}.

Specifically, small-loss tricks mean deep networks regard small-loss examples as ``clean'' examples, 
and only back-propagate such examples to update the model parameters. 
Mathematically, small-loss tricks are equivalent to constructing the restricted $\tilde{\ell}$,
where $\tilde{\ell} = \text{sort}(\ell,1-\tau)$. Namely, $\tilde{\ell}$ can be constructed by sorting $\ell$ from small to large, and fetching $1-\tau$ percentage of small loss ($\tau$ is noise rate).

\subsection{Self-training}

\begin{figure}[t]
\centering
\includegraphics[width=0.45\textwidth]{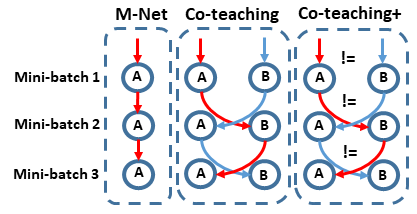}
\vspace{-10px}
\caption{Self-training (i.e., MentorNet, abbreviated as M-Net) vs. Co-training (i.e., Co-teaching and Co-teaching+).}
\label{net-structure}
\end{figure}

Based on small-loss tricks, the seminal works leverage self-training to improve the model robustness (left panel of
Fig.~\ref{net-structure}). There are two works as follows.

\subsubsection{MentorNet}
Jiang et al.~\cite{jiang2018mentornet} introduced MentorNet, which supervises the base deep network termed StudentNet. The key focus of MentorNet is to provide a curriculum for StudentNet. Instead of pre-defining, MentorNet learns a data-driven curriculum dynamically.

Mathematically, MentorNet $g_m$ can approximate a predefined curriculum via
minimizing the equation as follows:
\begin{equation*}
\arg\min_{\theta}
\sum\nolimits_{(x_i,y_i)\in \mathcal{D}}g_m(z_i;\theta)\ell_i + G(g_m(z_i;\theta);\lambda_1, \lambda_2),
\end{equation*}
where $\lambda_1, \lambda_2$ are regularization parameters. The first term denotes the curriculum-weighted loss and the second term
indicates the robust penalty $G$ (Eq.~(5) in \cite{jiang2018mentornet}). 
After optimizing the above objective, we can get
the closed-form solution as follows.
\begin{equation*}
g_m(z_i;\theta) 
\! = \! \left\{
\begin{array}{ll}
      \mathbf{1}(\ell_i \leq \lambda_1) & \text{if} \, \lambda_2 = 0,\\
      \min(\max(0, 1 - \nicefrac{\ell_i - \lambda_1}{\lambda_2}),1) & \text{if} \, \lambda_2 \neq 0.\\
\end{array}
\right.
\end{equation*}
Intuitively, when $\lambda_2 = 0$, MentorNet only provides small-loss samples with $\ell_i < \lambda_1$. When $\lambda_2 \neq 0$, MentorNet will not provide big-loss samples, namely, samples with loss larger than ($\lambda_1 + \lambda_2$) will not be selected during training. Meanwhile, MentorNet can also discover new curriculums from data directly, which is unrelated to small-loss tricks.

\subsubsection{Learning to Reweight}
Ren et al.~\cite{ren2018learning} employed a meta-learning framework to assign different weights to the training examples based on their gradient directions. In general,  small-loss examples are assigned  more weights, since small-loss examples are more likely to be clean. In general, they believed that the best example weighting should minimize the loss of a set of unbiased clean validation examples. Namely, they performed validation at every training iteration to dynamically determine the example weights of the current batch. Mathematically, they hoped to learn a reweighting of the inputs via minimizing a weighted loss:
\begin{equation}
\theta^*(w) = \arg\min_{\theta}
\sum\nolimits_{i=1}^N w_i \ell_i(\theta),
\end{equation}
where the training loss $\ell_i$ associates with a training set
$\{(x_i,y_i)\}_{i=1}^N$. Note that $w_i$ can be viewed as training hyperparameters, and the optimal selection of $w$ is based on its validation performance: 
\begin{equation}
w^* = \arg\min_{w}\nicefrac{1}{M}
\sum\nolimits_{i=1}^M \ell_i^v(\theta^*(w)),
\end{equation}
where the validation loss $\ell_i^v$ associates with a small validation set $\{(x_i^v,y_i^v)\}_{i=1}^M$. To realize ``Learning to Reweight'', there are three technical steps. First, they fed forward and backward noisy training examples via the training loss, which updates the model parameter $\theta$ and calculates $\nabla\theta$. Second, the $\nabla\theta$ affects the validation networks, where they fed forward and backward clean validation examples via the validation loss. Lastly, training networks leverage meta-learning to update example weights $w$ via backward on backward (i.e., taking a second-order gradient for the example weights). Note that the same strategy can also be  used for class-imbalance problems, in which big-loss tricks are preferred, since they are more likely to be the minority class~\cite{malisiewicz2011ensemble}.

\subsection{Co-training}

Although self-training works well, in the long term, the error will still accumulate, which motivates us to explore Co-training~\cite{blum1998combining} based methods (right panel of Fig.~\ref{net-structure}).

\subsubsection{Co-teaching/Co-teaching+}
Han et al.~\cite{han2018co} proposed a new deep learning paradigm called ``Co-teaching''. In general, instead of training a single deep network, they trained two deep networks simultaneously, and let them teach each other given every mini-batch. 
Specifically, each network feeds forward all data and selects some data with possibly clean labels; 
then, two networks communicate with each other what data in this mini-batch should be used for training; 
lastly, each network back-propagates the data selected by its peer network and updates itself. The selection criterion is still based on the small-loss trick.

In MentorNet~\cite{jiang2018mentornet}, the error from one network will be directly transferred back to itself in the second mini-batch of data, and the error should be increasingly accumulated. However, in Co-teaching, since two networks have different learning abilities, they can filter different types of error introduced by noisy labels. Namely, in this exchange procedure, the error flows can be reduced by peer networks mutually. However, with the increase of training epochs, two networks will converge to a consensus and Co-teaching will reduce to the self-training MentorNet in function. Note that the principle of ensemble learning is to keep different classifiers diverged~\cite{dietterich2002ensemble}.

Yu et al.~\cite{yu2019does} introduced the ``update by disagreement''~\cite{malach2017decoupling} strategy to keep Co-teaching diverged and named their method Co-teaching+. 
In general, Co-teaching+ consists of the disagreement-update step and cross-update step. In the disagreement-update step, two networks feed forward and predict all data first, and only keep prediction-disagreed data. This step indeed keeps two networks diverged. The cross-update step has been explored in Co-teaching. Note that both Co-teaching and Co-teaching+ share the same dropping rate for big-loss examples, which was hand-designed. Via both methods, we summarize three key factors in this line research: (1) using the small-loss trick; (2) cross-updating parameters of two networks; and (3) keeping two networks diverged.

\subsubsection{Beyond Co-teaching}

After 2018, 
there are several important works to further improve Co-teaching. 
Here, we specify two representative works based on Co-teaching and go beyond. 

\begin{itemize}[leftmargin=*]
\item 
Chen et al.~\cite{chen2019understanding} used cross-validation to randomly split noisy datasets, which identifies most samples that have correct labels. In general, they designed the Iterative Noisy Cross-Validation (INCV) method to select a subset of samples, which has a much smaller noise ratio than the original dataset. Then, they leveraged Co-teaching for further training over a selected subset. Apart from selecting clean samples, INCV removes samples that have large losses at each iteration.

\item 
Yao et al. \cite{yao2020s2e} used automated machine learning (AutoML) 
\cite{yao2018taking,automl_book}
to explore the memorization effect thus improve Co-teaching. 
It is noted that both Co-teaching and Co-teaching+ share the same dropping rate for big-loss examples, 
which was hand-designed. However, such a rate is critical in training deep networks. 
Specifically, Yao et al.~\cite{yao2020s2e} designed a domain-specific search and proposed
a novel Newton algorithm to solve the bi-level optimization problem efficiently. 
To explore the optimal rate $R(\cdot)$, they formulated the problem as
\begin{align*}
R^* & = \arg\min\nolimits_{R(\cdot) \in \mathcal{F}} \mathcal{L}_{\text{val}}(f(w^*;R),\mathcal{D}_{\text{val}})
\\
\text{s.t.} \;
& w^* = \arg\min\nolimits_{w} \mathcal{L}_{\text{tr}}(f(w^*;R),\mathcal{D}_{\text{tr}}),
\end{align*}
where $\mathcal{F}$ is the search space of $R(\cdot)$ exploring a general pattern of the memorization effect. 
Such a prior on $\mathcal{F}$ not only allows 
efficient bi-level optimization
but also boosts the final learning performance. 

\item
Motivated by MixMatch\cite{berthelot2019mixmatch},
Li et al.~\cite{li2020dividemix} promoted a novel framework termed DivideMix by leveraging semi-supervised learning techniques. 
In high-level, DivideMix used a Gaussian Mixture Model (GMM) to dynamically
divides the training data into two parts. The first part includes labeled data
with clean labels; while the second part includes unlabeled data with noisy
labels. During the semi-supervised learning phase, they leveraged variants of
Co-training, such as Co-refinement~\cite{li2020dividemix} on labeled data and
Co-guessing~\cite{li2020dividemix} on unlabeled data. Specifically, once the data is divided into labeled and unlabeled data, they conducted Co-refinement for labeled data, which linearly combines the ground-truth label with the network's prediction and sharpens the refined label. Then they conducted Co-guessing for unlabeled data, which averages the predictions from both networks. After Co-refinement and Co-guessing, they followed the routine MixMatch to mix the data, and updated the model parameters.
\end{itemize}

\subsection{Beyond Memorization}
This section focuses on solving LNRL by leveraging 
the unique benefits of deep models, especially for their memorization effects. However, besides memorization, there are two new branches based on deep models as follows.

\subsubsection{Pre-training}
In many CV and NLP applications, the pre-training paradigm has become a commonplace, especially when data is scarce in the target domain. 
Hendrycks et al.~\cite{hendrycks2019using} demonstrated that pre-training can improve model robustness and uncertainty, 
including adversarial robustness and label corruption.

Normally, pre-training is carried out on a bigger dataset first, 
and fine-tuning is then applied to the pre-trained model on a smaller target dataset. 
For example, if we design an LNRL method for image classification with label noise, we pre-train a model on ImageNet via an LNRL method first. 
Then, we fine-tune the pre-trained model on the target dataset via an LNRL method. 
Note that the pre-training approach has been demonstrated in many robustness and uncertainty scenarios~\cite{hendrycks2019using,liu2020energy},
including label noise, adversarial examples, class imbalance, out-of-distribution detection and calibration.

\subsubsection{Deep k-NN}
Bahri et al.~\cite{bahri2020deep} proposed the deep k-NN method, which works on an intermediate layer 
of a preliminary deep model $\mathcal{M}$ to filter mislabeled training data. In
high-level, deep k-NN filtering consists of two steps. In the first step, a model
$\mathcal{M}$ with architecture $\mathcal{A}$ is trained 
to filter noisy data $\mathcal{D}_{\text{noisy}}$ via the k-NN algorithm, which identifies and removes examples whose labels disagree with their neighbors. After filtering $\mathcal{D}_{\text{noisy}}$, 
in the second step, a final model with architecture $\mathcal{A}$ is re-trained on $\mathcal{D}_{\text{filtered}} \cup \mathcal{D}_{\text{clean}}$.

\section{Future Works}
\label{sec:fuworks}
Starting from 1988, label-noise learning has been investigated for more than three decades, evolving from statistical learning to representation learning. Especially from 2012, representation learning becomes increasingly important, which directly gives birth to the above LNRL methods. Similar to the other areas in machine learning, we hope to propose not only new methods, but also new research directions, which can broaden and boost  LNRL research in both academia and industry.

\subsection{Build up New Datasets}
In LNRL, the first thing we should do is to construct new datasets with real noise, which is critical to the rapid development of LNRL. To our best knowledge, 
most  researchers test their LNRL methods on simulated datasets, 
such as \textit{MNIST} and \textit{CIFAR-10}. To make further breakthroughs, we should build  new datasets with real noise, 
such as \textit{Clothing1M} \cite{xiao2015learning}.

Note that, 
similar to \textit{ImageNet}, 
many researchers train deep networks to overfit \textit{Clothing1M} via different tricks, 
which may not touch the core issues of LNRL.
This motivates us to rethink real datasets in LNRL. 5 years after the birth of Clothing1M, Jiang et al.~\cite{jiang2020cont} proposed a new but realistic type of noisy dataset called ``web-label noise'' (or red noise), which enables us to conduct controlled experiments systematically in realistic scenarios. Another interesting point is that benchmark datasets with real noise mainly focus on image classification, instead of natural language, speech processing, and various data collected from real sensors. Obviously, these directions also involve label noise, which need to be addressed further.

To sum up, we should build  new benchmark datasets with \textit{real noise}, not only for images but also for language, speech, and various sensor data. Normally, the better datasets can boost  rapid development of LNRL.

\subsection{Instance-dependent LNRL}
\label{sec:fu:idlnrl}

Previously in Section~\ref{sec:thm:idata},
we have seen that CCN is a popular assumption in LNRL.
However, the CCN model is only an approximation to the real-world noise, which may not always work well in practice.
To directly model  real-world noise, we should consider  features in the label corruption process. 
This motivates us to explore the instance-dependent noise (IDN) model, which is formulated as $p(\bar{Y}|Y,X)$~\cite{menon2018learning}. 
Specifically, the IDN model considers a more general noise, in which the probability that an instance is mislabeled depends on both its class and features. Intuitively, this noise is quite realistic, as poor-quality or ambiguous instances are more likely to be mislabeled in real-world datasets. However, it is much more complex to formulate the IDN model, since the probability of a mislabeled instance is a function of not only the label space but also the input space that can be very high-dimensional. Moreover, without some extra assumptions/information, IDN is unidentifiable~\cite{cheng2020learning}.

Towards the IDN model, there are several seminal explorations. For instance, Menon et al.~\cite{menon2018learning} proposed the boundary-consistent noise model, which considers stronger noise for samples closer to the decision boundary of the Bayes optimal classifier. However, this model is restricted to binary classification and cannot estimate noise functions. Cheng et al.~\cite{cheng2020learning} recently studied a particular case of the IDN model, in which the noise functions are upper-bounded. Nonetheless, their method is limited to binary classification and has only been tested on small datasets. Berthon et al.~\cite{berthon2020confidence} proposed to tackle the IDN model from the new perspective, by considering confidence scores to be available for the label of each instance. They term this new setting confidence-scored IDN (CSIDN). Based on CSIDN, they derived an instance-level forward correction algorithm. More recently, Xia et al.~\cite{xia2020part}, Cheng et al.~\cite{cheng2021learning} and Zhu et al.~\cite{zhu2020second} explored this direction in depth.

\subsection{Adversarial LNRL}
When discussing robustness, 
one may think about adversarial robustness~\cite{Goodfellow2015ExplainingAH}. 
However, adversarial robustness is obtained via adversarial training, in which the features are adversarially perturbed while the labels are clean. 
Is this the optimal way to formulate the adversarial robustness? In other words, would it be more useful to consider the scenario in which the features are adversarially perturbed while the labels are noisy? We term this adversarial LNRL.

Towards adversarial LNRL, there are two seminal works. For example, Wang et al.~\cite{wang2019improving} proposed a new defense algorithm called misclassification aware adversarial training (MART), which explicitly differentiates the misclassified examples (i.e., label noise) and correctly classified examples during  training. To address this issue, MART introduces misclassification aware regularization, namely $1/n\sum_{i=1}^{n}\mathbf{1}(h_{\theta}(x_i)\neq h_{\theta}(\hat{x}'_i))\cdot\mathbf{1}(h_{\theta}(x_i)\neq y_i)$, where $h_{\theta}$ denotes a DNN classifer with model parameter $\theta$, $(x_i, y_i) $ denotes the $i$th pair of features and label, and $\hat{x}'_i$ denotes the adversarial example generated by (5) in \cite{wang2019improving}. Intuitively, $\mathbf{1}(h_{\theta}(x_i)\neq y_i)$ denotes the misclassified examples, which can be closely connected with label noise. Meanwhile, Zhang et al.~\cite{zhang2020attacks} considered the same issue, namely misclassified examples in adversarial training. Specifically, they proposed friendly adversarial training (FAT), which trains deep networks using the wrongly-predicted adversarial data minimizing the loss and the correctly-predicted adversarial data maximizing the loss. To realize FAT, they introduced early-stopped PGD. Namely, once the adversarial data is misclassified by the current model, they  stopped the PGD iterations early. In high-level, the objective of FAT is min-min optimization instead of min-max optimization in standard adversarial training.

\subsection{Beyond Labels: Noisy Data}

We have envisioned three promising directions above, which belong to the vertical domain in LNRL. However, we hope to explore the horizontal domain more, namely, noisy data instead of only noisy labels. Here, we summarize different formats of noisy data, and show some preliminary works.

\begin{itemize}[leftmargin=*]
\item 
\textit{Feature}:
Naturally, label noise can arouse us to consider feature noise, where the adversarial example is one of the special cases in feature noise~\cite{Goodfellow2015ExplainingAH}. The problem of feature noise is formulated as $p(\bar{X}|Y)$, in which the features are corrupted but the labels are intact.
Therefore, adversarial training can be the main tool to defend from adversarial examples. Note that there exists another feature noise called random perturbation~\cite{fawzi2016robustness}. To address this issue, Zhang et al.~\cite{zhang2019towards} proposed a robust ResNet, which is motivated by dynamical systems. Specifically, they characterized ResNet based on an explicit Euler method. This allows us to exploit the step factor in the Euler method to control the robustness of ResNet. They  proved that a small step factor can benefit its training and generalization robustness during backward and forward propagation. Namely, controlling the step factor robustifies deep networks, which can alleviate feature noise.

\item \textit{Preference}:
Han et al.~\cite{han2018robust} and Pan et al.~\cite{pan2018stagewise} tried to address preference noise in ranking problems,
which plays an important role in our daily life, such as ordinal peer grading, online image-rating and online product recommendation.
Specifically, Han et al.~\cite{han2018robust} proposed the ROPAL model, which integrates the Plackett-Luce model with a denoising vector. Based on the Kendall-tau distance, this vector corrects $k$-ary noisy preferences with a certain probability. However, ROPAL cannot handle the dynamic length of $k$-ary noisy preferences, which motivated Pan et al.~\cite{pan2018stagewise} to propose COUPLE, which leverages stagewise learning to break the limit of fixed length. To update the parameters of both models, they used online Bayesian inference.

\item \textit{Domain}:
Domain adaptation (DA) is one of the fundamental problems in machine learning, when the data volume in the target domain is scarce. Previous DA methods assume that labeled data in the source domain is  clean. However, in practice, labeled data in the source domain may come from amateur annotators or the Internet due to its large volume. This issue brings us a new setting, where labels in the source domain are noisy. We call this setting \textit{wild domain adaptation (WDA)}.
There are two seminal works. Specifically, to handle WDA, Liu et al.~\cite{liu2019butterfly} proposed the Butterfly framework, 
which maintains four deep networks simultaneously. 
Butterfly can obtain high-quality domain-invariant representations (DIR) and target-specific representations (TSR) in an iterative manner.
Meanwhile, Yu et al.~\cite{yu2020rda} proposed a novel Denoising Conditional Invariant Component (DCIC) framework, 
which provably ensures extracting invariant representations and estimating the label distribution in the target domain with no bias.

\item \textit{Similarity}:
Similarity-based learning~\cite{bao2018classification} is one of the emerging weakly-supervised problems, where similar data pairs (i.e., two examples belonging to the same class) and unlabeled data are available. Compared to class labels, similarity labels are easier to obtain, especially for some sensitive issues, e.g., religion and politics. For example, for sensitive matters, people often hesitate to directly answer ``What is your opinion on issue A?''; while they are more likely to answer ``With whom do you share the same opinion on issue A?''. Therefore, similarity labels are easier to obtain. However, for some cases, people may not be willing to provide their real thoughts even facing easy questions. Therefore, noisy-similarity-labeled data are very challenging. Wu et al.~\cite{wu2020multi} employed a noise transition matrix to model similarity noise, which has been integrated into a deep learning system.

\item \textit{Graph}:
Graph neural networks (GNNs) are very hot in the machine learning community~\cite{hamilton2017representation}. 
However, are GNNs robust to noise? For example, once the node or edge is corrupted, the performance of GNNs will certainly deteriorate. Since GNNs are highly related to discrete and combinatorial optimization, LNRL methods cannot be directly deployed. Therefore, it is very meaningful to robustify GNNs under the node or edge noise, in which the noise can occur in labels and features. Recently, Wang et al.~\cite{wang2020crossgraph} proposed a robust and unsupervised embedding framework called Cross-Graph, which can handle structural corruption in attributed graphs. Meanwhile, since Hendrycks et al.~\cite{hendrycks2019using} discovered that pre-training can improve  model robustness and uncertainty, we may leverage strategies for pre-training GNNs~\cite{hu2019strategies} to overcome the issues of graph noise.

\item \textit{Demonstration}:
The goal of imitation learning (IL) is to learn a good policy from high-quality demonstrations~\cite{abbeel2004apprenticeship,le2016smooth}. However, the quality of demonstrations in reality should be diverse, since it is easier and cheaper to collect demonstrations from a mix of experts and amateurs. This brings us a new setting of IL called diverse-quality demonstrations, where low-quality demonstrations are highly noisy~\cite{mandlekar2018roboturk}. When experts provide additional information about the quality, learning from diverse-quality demonstrations becomes relatively easy, since the quality can be estimated by their confidence scores~\cite{wu2019imitation}, ranking scores~\cite{brown2019extrapolating} and a small number of high-quality demonstrations~\cite{audiffren2015maximum}. However, without the availability of experts, these methods may not work well. Recently, Tangkaratt et al.~\cite{tangkaratt2019vild} pushed forward this line, and proposed to model the quality with a probabilistic graphic model termed VILD. Specifically, they estimated the quality along with a reward function that represents an intention of experts' decision making. Moreover, they used a variational approach to handle large state-action spaces, and employed importance sampling to improve  data efficiency.
\end{itemize}

%
%

\section{Conclusions}
\label{sec:conclusions}

In this survey, 
we reviewed the history of label-noise representation learning (LNRL), 
and formally defined LNRL from the view of machine learning. 
Via the lens of representation learning theory and empirical experiments, we tried to understand the mechanism of deep networks under label noise. Based on the above analysis, 
we categorized LNRL methods into three perspectives, 
i.e., data, objective and model. 
Specifically, 
under such a taxonomy, we provided thorough discussion of the pros and cons across different categories.
Moreover, we summarized the essential components of robust LNRL, which can enlighten new directions in LNRL. 
Lastly, we proposed four possible research directions. The first three directions mainly focus on pushing the knowledge boundary of LNRL, including building up new datasets, instance-dependent LNRL and adversarial LNRL. The last direction is beyond LNRL, which learns from various types of noisy data, such as preference-noise, domain-noise, similarity-noise, graph-noise and demonstration-noise. Ultimately, we hope to uncover the secret of data-noise representation learning, and formulate a general framework in the near future.

\section*{Acknowledgments}

BH was supported by the RGC Early Career Scheme No. 22200720, NSFC Young Scientists Fund No. 62006202 and HKBU CSD Departmental Incentive Grant.
TLL was supported by Australian Research Council Project DE-190101473. IWT was supported by Australian Research Council under Grants DP180100106 and DP200101328. GN and MS were supported by JST AIP Acceleration Research Grant Number JPMJCR20U3, Japan. MS was also supported by the Institute for AI and Beyond, UTokyo.

\bibliographystyle{IEEEtran}
\bibliography{survey_paper}

\begin{thebibliography}{100}
\providecommand{\url}[1]{#1}
\csname url@samestyle\endcsname
\providecommand{\newblock}{\relax}
\providecommand{\bibinfo}[2]{#2}
\providecommand{\BIBentrySTDinterwordspacing}{\spaceskip=0pt\relax}
\providecommand{\BIBentryALTinterwordstretchfactor}{4}
\providecommand{\BIBentryALTinterwordspacing}{\spaceskip=\fontdimen2\font plus
\BIBentryALTinterwordstretchfactor\fontdimen3\font minus
  \fontdimen4\font\relax}
\providecommand{\BIBforeignlanguage}[2]{{%
\expandafter\ifx\csname l@#1\endcsname\relax
\typeout{** WARNING: IEEEtran.bst: No hyphenation pattern has been}%
\typeout{** loaded for the language `#1'. Using the pattern for}%
\typeout{** the default language instead.}%
\else
\language=\csname l@#1\endcsname
\fi
#2}}
\providecommand{\BIBdecl}{\relax}
\BIBdecl

\bibitem{angluin1988learning}
D.~Angluin and P.~Laird, ``Learning from noisy examples,'' \emph{ML}, vol.~2,
  no.~4, pp. 343--370, 1988.

\bibitem{krizhevsky2012imagenet}
A.~Krizhevsky, I.~Sutskever, and G.~E. Hinton, ``Imagenet classification with
  deep convolutional neural networks,'' in \emph{NeurIPS}, 2012, pp.
  1097--1105.

\bibitem{deng2009imagenet}
J.~Deng, W.~Dong, R.~Socher, L.-J. Li, K.~Li, and L.~Fei-Fei, ``Imagenet: A
  large-scale hierarchical image database,'' in \emph{CVPR}, 2009, pp.
  248--255.

\bibitem{reddy2019supervised}
C.~K. Reddy, R.~Cutler, and J.~Gehrke, ``Supervised classifiers for audio
  impairments with noisy labels,'' in \emph{INTERSPEECH}, 2019.

\bibitem{natarajan2013learning}
N.~Natarajan, I.~S. Dhillon, P.~K. Ravikumar, and A.~Tewari, ``Learning with
  noisy labels,'' in \emph{NeurIPS}, 2013.

\bibitem{jiang2018mentornet}
L.~Jiang, Z.~Zhou, T.~Leung, L.-J. Li, and L.~Fei-Fei, ``Mentornet: Learning
  data-driven curriculum for very deep neural networks on corrupted labels,''
  in \emph{ICML}, 2018, pp. 2304--2313.

\bibitem{han2018co}
B.~Han, Q.~Yao, X.~Yu, G.~Niu, M.~Xu, W.~Hu, I.~Tsang, and M.~Sugiyama,
  ``Co-teaching: Robust training of deep neural networks with extremely noisy
  labels,'' in \emph{NeurIPS}, 2018, pp. 8527--8537.

\bibitem{frenay2013classification}
B.~Fr{\'e}nay and M.~Verleysen, ``Classification in the presence of label
  noise: a survey,'' \emph{TNNLS}, vol.~25, no.~5, pp. 845--869, 2013.

\bibitem{algan2019image}
G.~Algan and I.~Ulusoy, ``Image classification with deep learning in the
  presence of noisy labels: A survey,'' \emph{arXiv preprint arXiv:1912.05170},
  2019.

\bibitem{karimi2020deep}
D.~Karimi, H.~Dou, S.~K. Warfield, and A.~Gholipour, ``Deep learning with noisy
  labels: exploring techniques and remedies in medical image analysis,''
  \emph{Medical Image Analysis}, 2020.

\bibitem{song2020learning}
H.~Song, M.~Kim, D.~Park, and J.-G. Lee, ``Learning from noisy labels with deep
  neural networks: A survey,'' \emph{arXiv preprint arXiv:2007.08199}, 2020.

\bibitem{lawrence2001estimating}
N.~Lawrence and B.~Sch{\"o}lkopf, ``Estimating a kernel fisher discriminant in
  the presence of label noise,'' in \emph{ICML}, 2001, pp. 306--306.

\bibitem{bartlett2006convexity}
P.~L. Bartlett, M.~I. Jordan, and J.~D. McAuliffe, ``Convexity, classification,
  and risk bounds,'' \emph{JASA}, vol. 101, no. 473, pp. 138--156, 2006.

\bibitem{crammer2006online}
K.~Crammer, O.~Dekel, J.~Keshet, S.~Shalev-Shwartz, and Y.~Singer, ``Online
  passive-aggressive algorithms,'' \emph{JMLR}, vol.~7, no. Mar, pp. 551--585,
  2006.

\bibitem{scott2013classification}
C.~Scott, G.~Blanchard, and G.~Handy, ``Classification with asymmetric label
  noise: Consistency and maximal denoising,'' in \emph{COLT}, 2013, pp.
  489--511.

\bibitem{van2015learning}
B.~van Rooyen, A.~Menon, and R.~Williamson, ``Learning with symmetric label
  noise: The importance of being unhinged,'' in \emph{NeurIPS}, 2015, pp.
  10--18.

\bibitem{liu2015classification}
T.~Liu and D.~Tao, ``Classification with noisy labels by importance
  reweighting,'' \emph{TPAMI}, vol.~38, no.~3, pp. 447--461, 2015.

\bibitem{sukhbaatar2014training}
S.~Sukhbaatar, J.~Bruna, M.~Paluri, L.~Bourdev, and R.~Fergus, ``Training
  convolutional networks with noisy labels,'' in \emph{ICLR Workshop}, 2015.

\bibitem{reed2014training}
S.~Reed, H.~Lee, D.~Anguelov, C.~Szegedy, D.~Erhan, and A.~Rabinovich,
  ``Training deep neural networks on noisy labels with bootstrapping,'' in
  \emph{ICLR Workshop}, 2015.

\bibitem{azadi2015auxiliary}
S.~Azadi, J.~Feng, S.~Jegelka, and T.~Darrell, ``Auxiliary image regularization
  for deep cnns with noisy labels,'' in \emph{ICLR}, 2016.

\bibitem{goldberger2016training}
J.~Goldberger and E.~Ben-Reuven, ``Training deep neural-networks using a noise
  adaptation layer,'' in \emph{ICLR}, 2017.

\bibitem{patrini2017making}
G.~Patrini, A.~Rozza, A.~Krishna~Menon, R.~Nock, and L.~Qu, ``Making deep
  neural networks robust to label noise: A loss correction approach,'' in
  \emph{CVPR}, 2017, pp. 1944--1952.

\bibitem{wang2017robust}
Y.~Wang, A.~Kucukelbir, and D.~M. Blei, ``Robust probabilistic modeling with
  bayesian data reweighting,'' in \emph{ICML}, 2017, pp. 3646--3655.

\bibitem{ren2018learning}
M.~Ren, W.~Zeng, B.~Yang, and R.~Urtasun, ``Learning to reweight examples for
  robust deep learning,'' in \emph{ICML}, 2018.

\bibitem{hendrycks2018using}
D.~Hendrycks, M.~Mazeika, D.~Wilson, and K.~Gimpel, ``Using trusted data to
  train deep networks on labels corrupted by severe noise,'' in \emph{NeurIPS},
  2018, pp. 10\,456--10\,465.

\bibitem{han2018masking}
B.~Han, J.~Yao, G.~Niu, M.~Zhou, I.~Tsang, Y.~Zhang, and M.~Sugiyama,
  ``Masking: A new perspective of noisy supervision,'' in \emph{NeurIPS}, 2018,
  pp. 5836--5846.

\bibitem{zhang2017mixup}
H.~Zhang, M.~Cisse, Y.~N. Dauphin, and D.~Lopez-Paz, ``mixup: Beyond empirical
  risk minimization,'' in \emph{ICLR}, 2018.

\bibitem{zhang2018generalized}
Z.~Zhang and M.~Sabuncu, ``Generalized cross entropy loss for training deep
  neural networks with noisy labels,'' in \emph{NeurIPS}, 2018, pp. 8778--8788.

\bibitem{ma2018dimensionality}
X.~Ma, Y.~Wang, M.~E. Houle, S.~Zhou, S.~M. Erfani, S.-T. Xia, S.~Wijewickrema,
  and J.~Bailey, ``Dimensionality-driven learning with noisy labels,'' in
  \emph{ICML}, 2018.

\bibitem{arazo2019unsupervised}
E.~Arazo, D.~Ortego, P.~Albert, N.~E. O'Connor, and K.~McGuinness,
  ``Unsupervised label noise modeling and loss correction,'' in \emph{ICML},
  2019.

\bibitem{hendrycks2019using}
D.~Hendrycks, K.~Lee, and M.~Mazeika, ``Using pre-training can improve model
  robustness and uncertainty,'' in \emph{ICML}, 2019.

\bibitem{charoenphakdee2019symmetric}
N.~Charoenphakdee, J.~Lee, and M.~Sugiyama, ``On symmetric losses for learning
  from corrupted labels,'' in \emph{ICML}, 2019.

\bibitem{thulasidasan2019combating}
S.~Thulasidasan, T.~Bhattacharya, J.~Bilmes, G.~Chennupati, and J.~Mohd-Yusof,
  ``Combating label noise in deep learning using abstention,'' in \emph{ICML},
  2019.

\bibitem{shu2019meta}
J.~Shu, Q.~Xie, L.~Yi, Q.~Zhao, S.~Zhou, Z.~Xu, and D.~Meng, ``Meta-weight-net:
  Learning an explicit mapping for sample weighting,'' in \emph{NeurIPS}, 2019,
  pp. 1919--1930.

\bibitem{menon2019can}
A.~K. Menon, A.~S. Rawat, S.~J. Reddi, and S.~Kumar, ``Can gradient clipping
  mitigate label noise?'' in \emph{ICLR}, 2020.

\bibitem{nguyen2019self}
D.~T. Nguyen, C.~K. Mummadi, T.~P.~N. Ngo, T.~H.~P. Nguyen, L.~Beggel, and
  T.~Brox, ``Self: Learning to filter noisy labels with self-ensembling,'' in
  \emph{ICLR}, 2020.

\bibitem{li2020dividemix}
J.~Li, R.~Socher, and S.~C. Hoi, ``Dividemix: Learning with noisy labels as
  semi-supervised learning,'' in \emph{ICLR}, 2020.

\bibitem{lyu2019curriculum}
Y.~Lyu and I.~W. Tsang, ``Curriculum loss: Robust learning and generalization
  against label corruption,'' in \emph{ICLR}, 2020.

\bibitem{han2020sigua}
B.~Han, G.~Niu, X.~Yu, Q.~Yao, M.~Xu, I.~Tsang, and M.~Sugiyama, ``Sigua:
  Forgetting may make learning with noisy labels more robust,'' in \emph{ICML},
  2020.

\bibitem{jiang2020cont}
L.~Jiang, D.~Huang, M.~Liu, and W.~Yang, ``Beyond synthetic noise: Deep
  learning on controlled noisy labels,'' in \emph{ICML}, 2020.

\bibitem{raghu2017expressive}
M.~Raghu, B.~Poole, J.~Kleinberg, S.~Ganguli, and J.~Sohl-Dickstein, ``On the
  expressive power of deep neural networks,'' in \emph{ICML}, 2017.

\bibitem{mitchell1997machine}
T.~Mitchell, \emph{Machine learning}, 1997.

\bibitem{mohri2018foundations}
M.~Mohri, A.~Rostamizadeh, and A.~Talwalkar, \emph{Foundations of machine
  learning}.\hskip 1em plus 0.5em minus 0.4em\relax MIT press, 2018.

\bibitem{Goodfellow2015ExplainingAH}
I.~J. Goodfellow, J.~Shlens, and C.~Szegedy, ``Explaining and harnessing
  adversarial examples,'' in \emph{ICLR}, 2015.

\bibitem{zhou2018brief}
Z.-H. Zhou, ``A brief introduction to weakly supervised learning,''
  \emph{National Science Review}, vol.~5, no.~1, pp. 44--53, 2018.

\bibitem{arpit2017closer}
D.~Arpit, S.~Jastrz{\k{e}}bski, N.~Ballas, D.~Krueger, E.~Bengio, M.~S. Kanwal,
  T.~Maharaj, A.~Fischer, A.~Courville, Y.~Bengio \emph{et~al.}, ``A closer
  look at memorization in deep networks,'' in \emph{ICML}, 2017.

\bibitem{Russakovsky2015ImageNetLS}
O.~Russakovsky, J.~Deng, H.~Su, J.~Krause, S.~Satheesh, S.~Ma, Z.~Huang,
  A.~Karpathy, A.~Khosla, M.~Bernstein, A.~Berg, and L.~Fei-Fei, ``Imagenet
  large scale visual recognition challenge,'' \emph{IJCV}, vol. 115, pp.
  211--252, 2015.

\bibitem{zhu2009introduction}
X.~Zhu and A.~B. Goldberg, ``Introduction to semi-supervised learning,''
  \emph{Synthesis lectures on artificial intelligence and machine learning},
  vol.~3, no.~1, pp. 1--130, 2009.

\bibitem{tarvainen2017mean}
A.~Tarvainen and H.~Valpola, ``Mean teachers are better role models:
  Weight-averaged consistency targets improve semi-supervised deep learning
  results,'' in \emph{NeurIPS}, 2017.

\bibitem{miyato2018virtual}
T.~Miyato, S.-i. Maeda, M.~Koyama, and S.~Ishii, ``Virtual adversarial
  training: a regularization method for supervised and semi-supervised
  learning,'' \emph{TPAMI}, vol.~41, no.~8, pp. 1979--1993, 2018.

\bibitem{berthelot2019mixmatch}
D.~Berthelot, N.~Carlini, I.~Goodfellow, N.~Papernot, A.~Oliver, and C.~A.
  Raffel, ``Mixmatch: A holistic approach to semi-supervised learning,'' in
  \emph{NeurIPS}, 2019.

\bibitem{elkan2008learning}
C.~Elkan and K.~Noto, ``Learning classifiers from only positive and unlabeled
  data,'' in \emph{KDD}, 2008.

\bibitem{kiryo2017positive}
R.~Kiryo, G.~Niu, M.~C. Du~Plessis, and M.~Sugiyama, ``Positive-unlabeled
  learning with non-negative risk estimator,'' in \emph{NeurIPS}, 2017.

\bibitem{hsieh2019classification}
Y.-G. Hsieh, G.~Niu, and M.~Sugiyama, ``Classification from positive, unlabeled
  and biased negative data,'' in \emph{ICML}, 2019.

\bibitem{ishida2018binary}
T.~Ishida, G.~Niu, and M.~Sugiyama, ``Binary classification from
  positive-confidence data,'' in \emph{NeurIPS}, 2018.

\bibitem{ishida2017learning}
T.~Ishida, G.~Niu, W.~Hu, and M.~Sugiyama, ``Learning from complementary
  labels,'' in \emph{NeurIPS}, 2017.

\bibitem{yu2018learning}
X.~Yu, T.~Liu, M.~Gong, and D.~Tao, ``Learning with biased complementary
  labels,'' in \emph{ECCV}, 2018.

\bibitem{ishida2019complementary}
T.~Ishida, G.~Niu, A.~Menon, and M.~Sugiyama, ``Complementary-label learning
  for arbitrary losses and models,'' in \emph{ICML}, 2019.

\bibitem{feng2019learning}
L.~Feng, T.~Kaneko, B.~Han, G.~Niu, B.~An, and M.~Sugiyama, ``Learning with
  multiple complementary labels,'' in \emph{ICML}, 2020.

\bibitem{du2013clustering}
M.~du~Plessis, G.~Niu, and M.~Sugiyama, ``Clustering unclustered data:
  Unsupervised binary labeling of two datasets having different class
  balances,'' in \emph{TAAI}, 2013.

\bibitem{lu2018minimal}
N.~Lu, G.~Niu, A.~Menon, and M.~Sugiyama, ``On the minimal supervision for
  training any binary classifier from only unlabeled data,'' in \emph{ICLR},
  2019.

\bibitem{lu2020mitigating}
N.~Lu, T.~Zhang, G.~Niu, and M.~Sugiyama, ``Mitigating overfitting in
  supervised classification from two unlabeled datasets: A consistent risk
  correction approach,'' in \emph{AISTATS}, 2020.

\bibitem{menon2018learning}
A.~Menon, B.~Van~Rooyen, and N.~Natarajan, ``Learning from binary labels with
  instance-dependent corruption,'' \emph{ML}, vol. 107, p. 1561–1595, 2018.

\bibitem{cheng2020learning}
J.~Cheng, T.~Liu, K.~Ramamohanarao, and D.~Tao, ``Learning with bounded
  instance-and label-dependent label noise,'' in \emph{ICML}, 2020.

\bibitem{xia2019anchor}
X.~Xia, T.~Liu, N.~Wang, B.~Han, C.~Gong, G.~Niu, and M.~Sugiyama, ``Are anchor
  points really indispensable in label-noise learning?'' in \emph{NeurIPS},
  2019.

\bibitem{patrini2016loss}
G.~Patrini, F.~Nielsen, R.~Nock, and M.~Carioni, ``Loss factorization, weakly
  supervised learning and label noise robustness,'' in \emph{ICML}, 2016, pp.
  708--717.

\bibitem{golowich2018size}
N.~Golowich, A.~Rakhlin, and O.~Shamir, ``Size-independent sample complexity of
  neural networks,'' in \emph{COLT}, 2018.

\bibitem{anthony2009neural}
M.~Anthony and P.~Bartlett, \emph{Neural network learning: Theoretical
  foundations}.\hskip 1em plus 0.5em minus 0.4em\relax Cambridge University
  Press, 2009.

\bibitem{li2020gradient}
M.~Li, M.~Soltanolkotabi, and S.~Oymak, ``Gradient descent with early stopping
  is provably robust to label noise for overparameterized neural networks,'' in
  \emph{AISTATS}, 2020.

\bibitem{zhang2016understanding}
C.~Zhang, S.~Bengio, M.~Hardt, B.~Recht, and O.~Vinyals, ``Understanding deep
  learning requires rethinking generalization,'' in \emph{ICLR}, 2017.

\bibitem{van2017theory}
B.~van Rooyen and R.~C. Williamson, ``A theory of learning with corrupted
  labels,'' \emph{JMLR}, vol.~18, no.~1, pp. 8501--8550, 2017.

\bibitem{xiao2015learning}
T.~Xiao, T.~Xia, Y.~Yang, C.~Huang, and X.~Wang, ``Learning from massive noisy
  labeled data for image classification,'' in \emph{CVPR}, 2015, pp.
  2691--2699.

\bibitem{dempster1977maximum}
A.~Dempster, N.~Laird, and D.~Rubin, ``Maximum likelihood from incomplete data
  via the em algorithm,'' \emph{Journal of the Royal Statistical Society:
  Series B}, vol.~39, no.~1, pp. 1--22, 1977.

\bibitem{szegedy2016rethinking}
C.~Szegedy, V.~Vanhoucke, S.~Ioffe, J.~Shlens, and Z.~Wojna, ``Rethinking the
  inception architecture for computer vision,'' in \emph{CVPR}, 2016.

\bibitem{lukasik2020does}
M.~Lukasik, S.~Bhojanapalli, A.~K. Menon, and S.~Kumar, ``Does label smoothing
  mitigate label noise?'' in \emph{ICML}, 2020.

\bibitem{frodesen1979probability}
A.~G. Frodesen, O.~Skjeggestad, and H.~Toefte, ``Probability and statistics in
  particle physics,'' 1979.

\bibitem{wasserman2013all}
L.~Wasserman, \emph{All of statistics: a concise course in statistical
  inference}.\hskip 1em plus 0.5em minus 0.4em\relax Springer Science \&
  Business Media, 2013.

\bibitem{wainwright2008graphical}
M.~Wainwright and M.~Jordan, ``Graphical models, exponential families, and
  variational inference, ser,'' \emph{Foundations and Trends in Machine
  Learning}, vol.~1, 2008.

\bibitem{grandvalet2005semi}
Y.~Grandvalet and Y.~Bengio, ``Semi-supervised learning by entropy
  minimization,'' in \emph{NeurIPS}, 2005.

\bibitem{lee2013pseudo}
D.-H. Lee, ``Pseudo-label: The simple and efficient semi-supervised learning
  method for deep neural networks,'' in \emph{ICML Workshop}, 2013.

\bibitem{chapelle2000vicinal}
O.~Chapelle, J.~Weston, L.~Bottou, and V.~Vapnik, ``Vicinal risk
  minimization,'' in \emph{NeurIPS}, 2000.

\bibitem{yao2020s2e}
Q.~Yao, H.~Yang, B.~Han, G.~Niu, and J.~T. Kwok, ``Searching to exploit
  memorization effect in learning with noisy labels,'' in \emph{ICML}, 2020.

\bibitem{gretton2009covariate}
A.~Gretton, A.~Smola, J.~Huang, M.~Schmittfull, K.~Borgwardt, and
  B.~Sch{\"o}lkopf, ``Covariate shift by kernel mean matching,'' \emph{Dataset
  shift in machine learning}, vol.~3, no.~4, p.~5, 2009.

\bibitem{bengio2009curriculum}
Y.~Bengio, J.~Louradour, R.~Collobert, and J.~Weston, ``Curriculum learning,''
  in \emph{ICML}, 2009.

\bibitem{wang2018iterative}
Y.~Wang, W.~Liu, X.~Ma, J.~Bailey, H.~Zha, L.~Song, and S.-T. Xia, ``Iterative
  learning with open-set noisy labels,'' in \emph{CVPR}, 2018.

\bibitem{laine2016temporal}
S.~Laine and T.~Aila, ``Temporal ensembling for semi-supervised learning,'' in
  \emph{ICLR}, 2017.

\bibitem{yu2019does}
X.~Yu, B.~Han, J.~Yao, G.~Niu, I.~W. Tsang, and M.~Sugiyama, ``How does
  disagreement help generalization against label corruption?'' in \emph{ICML},
  2019.

\bibitem{chen2019understanding}
P.~Chen, B.~Liao, G.~Chen, and S.~Zhang, ``Understanding and utilizing deep
  neural networks trained with noisy labels,'' in \emph{ICML}, 2019.

\bibitem{malisiewicz2011ensemble}
T.~Malisiewicz, A.~Gupta, and A.~Efros, ``Ensemble of exemplar-svms for object
  detection and beyond,'' in \emph{ICCV}, 2011.

\bibitem{blum1998combining}
A.~Blum and T.~Mitchell, ``Combining labeled and unlabeled data with
  co-training,'' in \emph{COLT}.

\bibitem{dietterich2002ensemble}
T.~Dietterich \emph{et~al.}, ``Ensemble learning,'' \emph{The handbook of brain
  theory and neural networks}, vol.~2, pp. 110--125, 2002.

\bibitem{malach2017decoupling}
E.~Malach and S.~Shalev-Shwartz, ``Decoupling" when to update" from" how to
  update",'' in \emph{NeurIPS}, 2017.

\bibitem{yao2018taking}
Q.~Yao and M.~Wang, ``Taking human out of learning applications: A survey on
  automated machine learning,'' arXiv preprint arXiv:1810.13306, Tech. Rep.,
  2018.

\bibitem{automl_book}
F.~Hutter, L.~Kotthoff, and J.~Vanschoren, Eds., \emph{Automated Machine
  Learning: Methods, Systems, Challenges}.\hskip 1em plus 0.5em minus
  0.4em\relax Springer, 2018.

\bibitem{liu2020energy}
W.~Liu, X.~Wang, J.~Owens, and Y.~Li, ``Energy-based out-of-distribution
  detection,'' \emph{NeurIPS}, 2020.

\bibitem{bahri2020deep}
D.~Bahri, H.~Jiang, and M.~Gupta, ``Deep k-nn for noisy labels,'' in
  \emph{ICML}, 2020.

\bibitem{berthon2020confidence}
A.~Berthon, B.~Han, G.~Niu, T.~Liu, and M.~Sugiyama, ``Confidence scores make
  instance-dependent label-noise learning possible,'' \emph{arXiv preprint
  arXiv:2001.03772}, 2020.

\bibitem{xia2020part}
X.~Xia, T.~Liu, B.~Han, N.~Wang, M.~Gong, H.~Liu, G.~Niu, D.~Tao, and
  M.~Sugiyama, ``Part-dependent label noise: Towards instance-dependent label
  noise,'' in \emph{NeurIPS}, 2020.

\bibitem{cheng2021learning}
H.~Cheng, Z.~Zhu, X.~Li, Y.~Gong, X.~Sun, and Y.~Liu, ``Learning with
  instance-dependent label noise: A sample sieve approach,'' in \emph{ICLR},
  2021.

\bibitem{zhu2020second}
Z.~Zhu, T.~Liu, and Y.~Liu, ``A second-order approach to learning with
  instance-dependent label noise,'' \emph{arXiv preprint arXiv:2012.11854},
  2020.

\bibitem{wang2019improving}
Y.~Wang, D.~Zou, J.~Yi, J.~Bailey, X.~Ma, and Q.~Gu, ``Improving adversarial
  robustness requires revisiting misclassified examples,'' in \emph{ICLR},
  2020.

\bibitem{zhang2020attacks}
J.~Zhang, X.~Xu, B.~Han, G.~Niu, L.~Cui, M.~Sugiyama, and M.~Kankanhalli,
  ``Attacks which do not kill training make adversarial learning stronger,'' in
  \emph{ICML}, 2020.

\bibitem{fawzi2016robustness}
A.~Fawzi, S.~Moosavi-Dezfooli, and P.~Frossard, ``Robustness of classifiers:
  from adversarial to random noise,'' in \emph{NeurIPS}, 2016.

\bibitem{zhang2019towards}
J.~Zhang, B.~Han, L.~Wynter, K.~H. Low, and M.~Kankanhalli, ``Towards robust
  resnet: A small step but a giant leap,'' in \emph{IJCAI}, 2019.

\bibitem{han2018robust}
B.~Han, Y.~Pan, and I.~W. Tsang, ``Robust plackett--luce model for k-ary
  crowdsourced preferences,'' \emph{ML}, vol. 107, no.~4, pp. 675--702, 2018.

\bibitem{pan2018stagewise}
Y.~Pan, B.~Han, and I.~W. Tsang, ``Stagewise learning for noisy k-ary
  preferences,'' \emph{ML}, vol. 107, no. 8-10, pp. 1333--1361, 2018.

\bibitem{liu2019butterfly}
F.~Liu, J.~Lu, B.~Han, G.~Niu, G.~Zhang, and M.~Sugiyama, ``Butterfly: A
  panacea for all difficulties in wildly unsupervised domain adaptation,''
  \emph{arXiv preprint arXiv:1905.07720}, 2019.

\bibitem{yu2020rda}
X.~Yu, T.~Liu, M.~Gong, K.~Zhang, K.~Batmanghelich, and D.~Tao, ``Label-noise
  robust domain adaptation,'' in \emph{ICML}, 2020.

\bibitem{bao2018classification}
H.~Bao, G.~Niu, and M.~Sugiyama, ``Classification from pairwise similarity and
  unlabeled data,'' in \emph{ICML}, 2018.

\bibitem{wu2020multi}
S.~Wu, X.~Xia, T.~Liu, B.~Han, M.~Gong, N.~Wang, H.~Liu, and G.~Niu,
  ``Multi-class classification from noisy-similarity-labeled data,''
  \emph{arXiv preprint arXiv:2002.06508}, 2020.

\bibitem{hamilton2017representation}
W.~L. Hamilton, R.~Ying, and J.~Leskovec, ``Representation learning on graphs:
  Methods and applications,'' \emph{IEEE Data Engineering Bulletin}, 2017.

\bibitem{wang2020crossgraph}
C.~Wang, B.~Han, S.~Pan, J.~Jiang, G.~Niu, and G.~Long, ``Cross-graph: Robust
  and unsupervised embedding for attributed graphs with corrupted structure,''
  in \emph{ICDM}, 2020.

\bibitem{hu2019strategies}
W.~Hu, B.~Liu, J.~Gomes, M.~Zitnik, P.~Liang, V.~Pande, and J.~Leskovec,
  ``Strategies for pre-training graph neural networks,'' in \emph{ICLR}, 2019.

\bibitem{abbeel2004apprenticeship}
P.~Abbeel and A.~Y. Ng, ``Apprenticeship learning via inverse reinforcement
  learning,'' in \emph{ICML}, 2004.

\bibitem{le2016smooth}
H.~Le, A.~Kang, Y.~Yue, and P.~Carr, ``Smooth imitation learning for online
  sequence prediction,'' in \emph{ICML}, 2016.

\bibitem{mandlekar2018roboturk}
A.~Mandlekar, Y.~Zhu, A.~Garg, J.~Booher, M.~Spero, A.~Tung, J.~Gao, J.~Emmons,
  A.~Gupta, E.~Orbay \emph{et~al.}, ``Roboturk: A crowdsourcing platform for
  robotic skill learning through imitation,'' in \emph{CoRL}, 2018.

\bibitem{wu2019imitation}
Y.-H. Wu, N.~Charoenphakdee, H.~Bao, V.~Tangkaratt, and M.~Sugiyama,
  ``Imitation learning from imperfect demonstration,'' in \emph{ICML}, 2019.

\bibitem{brown2019extrapolating}
D.~S. Brown, W.~Goo, P.~Nagarajan, and S.~Niekum, ``Extrapolating beyond
  suboptimal demonstrations via inverse reinforcement learning from
  observations,'' in \emph{ICML}, 2019.

\bibitem{audiffren2015maximum}
J.~Audiffren, M.~Valko, A.~Lazaric, and M.~Ghavamzadeh, ``Maximum entropy
  semi-supervised inverse reinforcement learning,'' in \emph{IJCAI}, 2015.

\bibitem{tangkaratt2019vild}
V.~Tangkaratt, B.~Han, M.~E. Khan, and M.~Sugiyama, ``Variational imitation
  learning with diverse-quality demonstrations,'' in \emph{ICML}, 2020.

\bibitem{bylander1994learning}
T.~Bylander, ``Learning linear threshold functions in the presence of
  classification noise,'' in \emph{COLT}, 1994, pp. 340--347.

\bibitem{dredze2008confidence}
M.~Dredze, K.~Crammer, and F.~Pereira, ``Confidence-weighted linear
  classification,'' in \emph{ICML}, 2008, pp. 264--271.

\bibitem{freund2009more}
Y.~Freund, ``A more robust boosting algorithm,'' \emph{arXiv preprint
  arXiv:0905.2138}, 2009.

\bibitem{cesa2011online}
N.~Cesa-Bianchi, S.~Shalev-Shwartz, and O.~Shamir, ``Online learning of noisy
  data,'' \emph{TIT}, vol.~57, no.~12, pp. 7907--7931, 2011.

\bibitem{misra2016seeing}
I.~Misra, C.~Lawrence~Zitnick, M.~Mitchell, and R.~Girshick, ``Seeing through
  the human reporting bias: Visual classifiers from noisy human-centric
  labels,'' in \emph{CVPR}, 2016.

\bibitem{krause2016unreasonable}
J.~Krause, B.~Sapp, A.~Howard, H.~Zhou, A.~Toshev, T.~Duerig, J.~Philbin, and
  L.~Fei-Fei, ``The unreasonable effectiveness of noisy data for fine-grained
  recognition,'' in \emph{ECCV}, 2016.

\bibitem{li2017learning}
Y.~Li, J.~Yang, Y.~Song, L.~Cao, J.~Luo, and L.-J. Li, ``Learning from noisy
  labels with distillation,'' in \emph{ICCV}, 2017.

\bibitem{northcutt2017learning}
C.~G. Northcutt, T.~Wu, and I.~L. Chuang, ``Learning with confident examples:
  Rank pruning for robust classification with noisy labels,'' in \emph{UAI},
  2017.

\bibitem{kim2019nlnl}
Y.~Kim, J.~Yim, J.~Yun, and J.~Kim, ``Nlnl: Negative learning for noisy
  labels,'' in \emph{ICCV}, 2019.

\bibitem{seo2019combinatorial}
P.~H. Seo, G.~Kim, and B.~Han, ``Combinatorial inference against label noise,''
  in \emph{NeurIPS}, 2019.

\bibitem{kaneko2019label}
T.~Kaneko, Y.~Ushiku, and T.~Harada, ``Label-noise robust generative
  adversarial networks,'' in \emph{CVPR}, 2019.

\bibitem{lamy2019noise}
A.~Lamy, Z.~Zhong, A.~K. Menon, and N.~Verma, ``Noise-tolerant fair
  classification,'' in \emph{NeurIPS}, 2019.

\bibitem{yao2019safeguarded}
J.~Yao, H.~Wu, Y.~Zhang, I.~W. Tsang, and J.~Sun, ``Safeguarded dynamic label
  regression for noisy supervision,'' in \emph{AAAI}, 2019.

\bibitem{branson2017lean}
S.~Branson, G.~Van~Horn, and P.~Perona, ``Lean crowdsourcing: Combining humans
  and machines in an online system,'' in \emph{CVPR}, 2017.

\bibitem{vahdat2017toward}
A.~Vahdat, ``Toward robustness against label noise in training deep
  discriminative neural networks,'' in \emph{NeurIPS}, 2017.

\bibitem{chang2017active}
H.-S. Chang, E.~Learned-Miller, and A.~McCallum, ``Active bias: Training more
  accurate neural networks by emphasizing high variance samples,'' in
  \emph{NeurIPS}, 2017.

\bibitem{khetan2017learning}
A.~Khetan, Z.~C. Lipton, and A.~Anandkumar, ``Learning from noisy
  singly-labeled data,'' in \emph{ICLR}, 2018.

\bibitem{tanaka2018joint}
D.~Tanaka, D.~Ikami, T.~Yamasaki, and K.~Aizawa, ``Joint optimization framework
  for learning with noisy labels,'' in \emph{CVPR}, 2018.

\bibitem{jenni2018deep}
S.~Jenni and P.~Favaro, ``Deep bilevel learning,'' in \emph{ECCV}, 2018.

\bibitem{wang2019symmetric}
Y.~Wang, X.~Ma, Z.~Chen, Y.~Luo, J.~Yi, and J.~Bailey, ``Symmetric cross
  entropy for robust learning with noisy labels,'' in \emph{ICCV}, 2019.

\bibitem{li2019learning}
J.~Li, Y.~Song, J.~Zhu, L.~Cheng, Y.~Su, L.~Ye, P.~Yuan, and S.~Han, ``Learning
  from large-scale noisy web data with ubiquitous reweighting for image
  classification,'' \emph{TPAMI}, 2019.

\bibitem{xu2019l_dmi}
Y.~Xu, P.~Cao, Y.~Kong, and Y.~Wang, ``L\_dmi: A novel information-theoretic
  loss function for training deep nets robust to label noise,'' in
  \emph{NeurIPS}, 2019.

\bibitem{liu2019peer}
Y.~Liu and H.~Guo, ``Peer loss functions: Learning from noisy labels without
  knowing noise rates,'' in \emph{ICML}, 2020.

\bibitem{ma2020normalized}
X.~Ma, H.~Huang, Y.~Wang, S.~Romano, S.~Erfani, and J.~Bailey, ``Normalized
  loss functions for deep learning with noisy labels,'' 2020.

\bibitem{veit2017learning}
A.~Veit, N.~Alldrin, G.~Chechik, I.~Krasin, A.~Gupta, and S.~Belongie,
  ``Learning from noisy large-scale datasets with minimal supervision,'' in
  \emph{CVPR}, 2017.

\bibitem{zhuang2017attend}
B.~Zhuang, L.~Liu, Y.~Li, C.~Shen, and I.~Reid, ``Attend in groups: a
  weakly-supervised deep learning framework for learning from web data,'' in
  \emph{CVPR}, 2017.

\bibitem{lee2018cleannet}
K.-H. Lee, X.~He, L.~Zhang, and L.~Yang, ``Cleannet: Transfer learning for
  scalable image classifier training with label noise,'' in \emph{CVPR}, 2018,
  pp. 5447--5456.

\bibitem{guo2018curriculumnet}
S.~Guo, W.~Huang, H.~Zhang, C.~Zhuang, D.~Dong, M.~R. Scott, and D.~Huang,
  ``Curriculumnet: Weakly supervised learning from large-scale web images,'' in
  \emph{ECCV}, 2018.

\bibitem{deng2019arcface}
J.~Deng, J.~Guo, N.~Xue, and S.~Zafeiriou, ``Arcface: Additive angular margin
  loss for deep face recognition,'' in \emph{CVPR}, 2019.

\bibitem{wang2019co}
X.~Wang, S.~Wang, J.~Wang, H.~Shi, and T.~Mei, ``Co-mining: Deep face
  recognition with noisy labels,'' in \emph{Proceedings of the IEEE
  international conference on computer vision}, 2019.

\bibitem{huang2019o2u}
J.~Huang, L.~Qu, R.~Jia, and B.~Zhao, ``O2u-net: A simple noisy label detection
  approach for deep neural networks,'' in \emph{Proceedings of the IEEE
  International Conference on Computer Vision}, 2019.

\bibitem{han2019deep}
J.~Han, P.~Luo, and X.~Wang, ``Deep self-learning from noisy labels,'' in
  \emph{ICCV}, 2019.

\bibitem{harutyunyan2020improving}
H.~Harutyunyan, K.~Reing, G.~V. Steeg, and A.~Galstyan, ``Improving
  generalization by controlling label-noise information in neural network
  weights,'' in \emph{ICML}, 2020.

\bibitem{wei2020combating}
H.~Wei, L.~Feng, X.~Chen, and B.~An, ``Combating noisy labels by agreement: A
  joint training method with co-regularization,'' in \emph{CVPR}, 2020.

\bibitem{zhang2020distilling}
Z.~Zhang, H.~Zhang, S.~O. Arik, H.~Lee, and T.~Pfister, ``Distilling effective
  supervision from severe label noise,'' in \emph{CVPR}, 2020.

\end{thebibliography}

\begin{IEEEbiography}[{\includegraphics[width = 1\textwidth,height = 0.12\textheight]{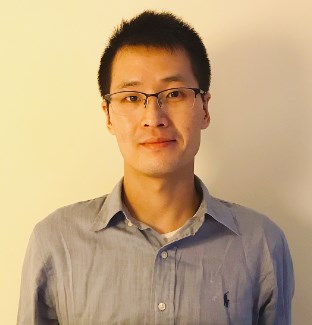}}]{Bo Han}
is an Assistant Professor of Computer Science at Hong Kong Baptist University, and a Visiting Scientist at RIKEN Center for Advanced Intelligence Project (RIKEN AIP). He was a Postdoc Fellow at RIKEN AIP (2019-2020). He received his Ph.D. degree in Computer Science from University of Technology Sydney in 2019. He has served as area chairs of NeurIPS'20 and ICLR'21. He received the RIKEN BAIHO Award (2019) and RGC Early Career Scheme (2020).
\end{IEEEbiography}

\begin{IEEEbiography}[{\includegraphics[width = 1\textwidth]{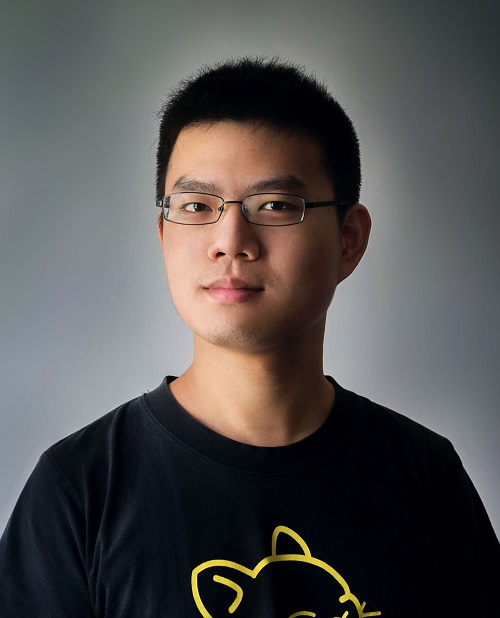}}]{Quanming Yao}
is a senior scientist in 4Paradigm. He obtained his Ph.D. degree at the Department of Computer Science and Engineering of Hong Kong University of Science and Technology (HKUST) in 2018 and received his bachelor degree at HuaZhong University of Science and Technology (HUST) in 2013. He has served as area chair of IJCAI'21. He is a receipt of Wunwen Jun Prize of Excellence Youth of Artificial Intelligence (issued by CAAI) and a winner of Google Fellowship (in machine learning).
\end{IEEEbiography}

\begin{IEEEbiography}[{\includegraphics[width = 1\textwidth]{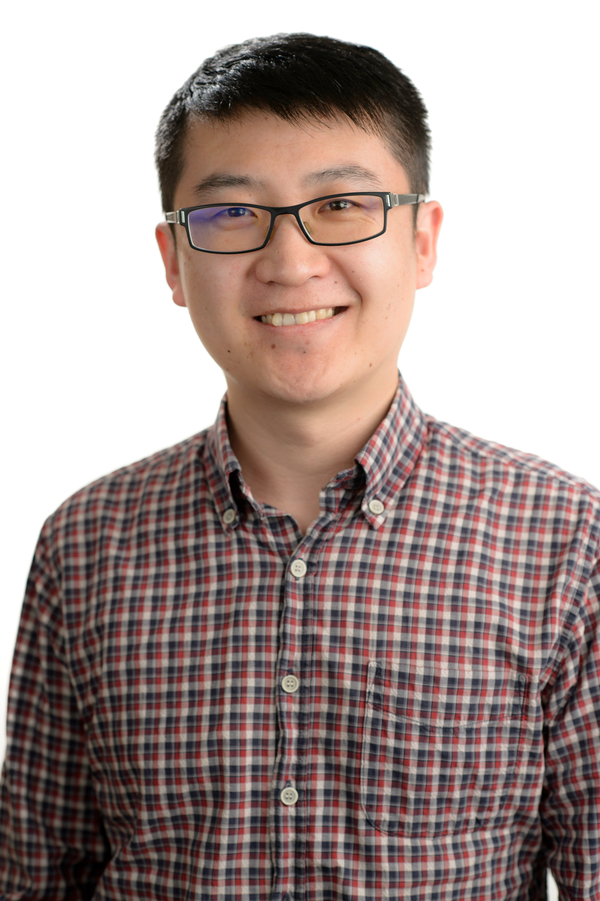}}]{Tongliang Liu} is a Lecturer (Assistant Professor) with the School of Computer Science at the University of Sydney. He is also a Visiting Scientist at RIKEN AIP. He has served as area chair of IJCAI'21. He is a recipient of Discovery Early Career Researcher Award (DECRA) from Australian Research Council (ARC) and was shortlisted for the J. G. Russell Award by Australian Academy of Science (AAS) in 2019.
\end{IEEEbiography}

\begin{IEEEbiography}[{\includegraphics[width = 1\textwidth]{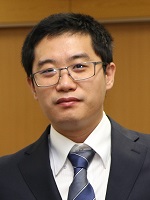}}]{Gang Niu}
is currently a research scientist (indefinite-term) at RIKEN Center for Advanced Intelligence Project.
He received the PhD degree in computer science from Tokyo Institute of Technology.
Before joining RIKEN as a research scientist, he was a senior software engineer at Baidu and then an assistant professor at the University of Tokyo. He has served as an area chair 10 times, including AISTATS'19, ICML'19-20, NeurIPS'19-20 and ICLR'21. He received the RIKEN BAIHO Award (2018).
\end{IEEEbiography}

\begin{IEEEbiography}[{\includegraphics[width = 1\textwidth, height = 0.14\textheight]{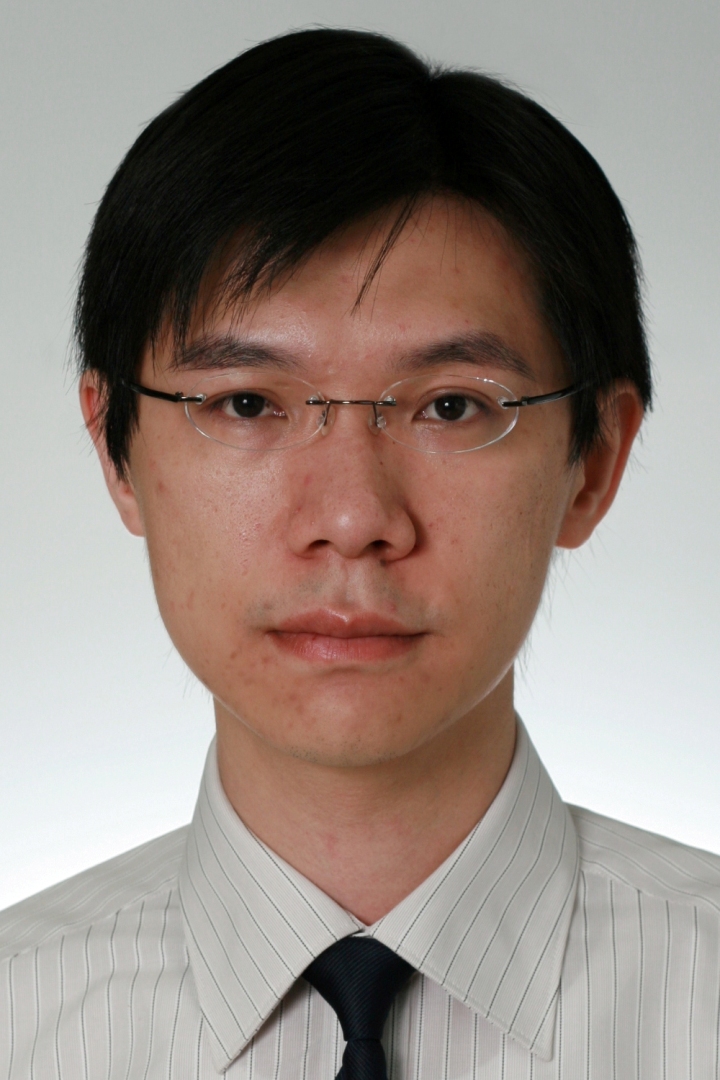}}]{Ivor W. Tsang} is Professor of Artificial Intelligence, at University of Technology Sydney. He is also the Research Director of the Australian Artificial Intelligence Institute. Prof. Tsang serves as a Senior Area Chair for Neural Information Processing Systems and Area Chair for International Conference on Machine Learning, and the Editorial Board for Journal of Machine Learning Research, Machine Learning, Journal of Artificial Intelligence Research and IEEE Transactions on Pattern Analysis and Machine Intelligence. He received the Australian Research Council Future Fellowship, and the AI 2000 AAAI/IJCAI Most Influential Scholar in Australia.
\end{IEEEbiography}

\begin{IEEEbiography}[{\includegraphics[width = 1\textwidth]{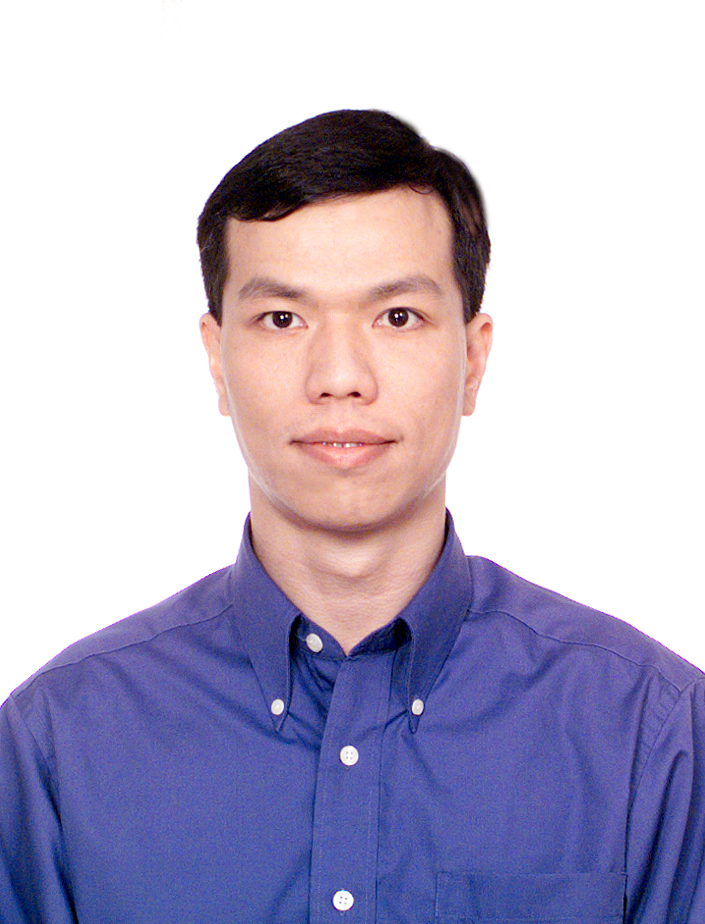}}]{James T. Kwok} (F’17) received the PhD degree in computer science from the Hong Kong University of Science and Technology, in 1996. He was
with the Department of Computer Science, Hong Kong Baptist University, Hong Kong, as an assistant professor. He is currently a professor with the Department of Computer Science and Engineering, Hong Kong University of Science and Technology. He has been a program cochair for a number of international conferences, and served as an associate editor for the IEEE Transactions on Neural Networks and Learning Systems and Artificial Intelligence Journal. He has also served as (Senior) Area Chairs of NeurIPS, ICML, ICLR, IJCAI, AAAI and ECML. He is a fellow of the IEEE.
\end{IEEEbiography}
\vspace{-20em}

\begin{IEEEbiography}[{\includegraphics[width = 1\textwidth]{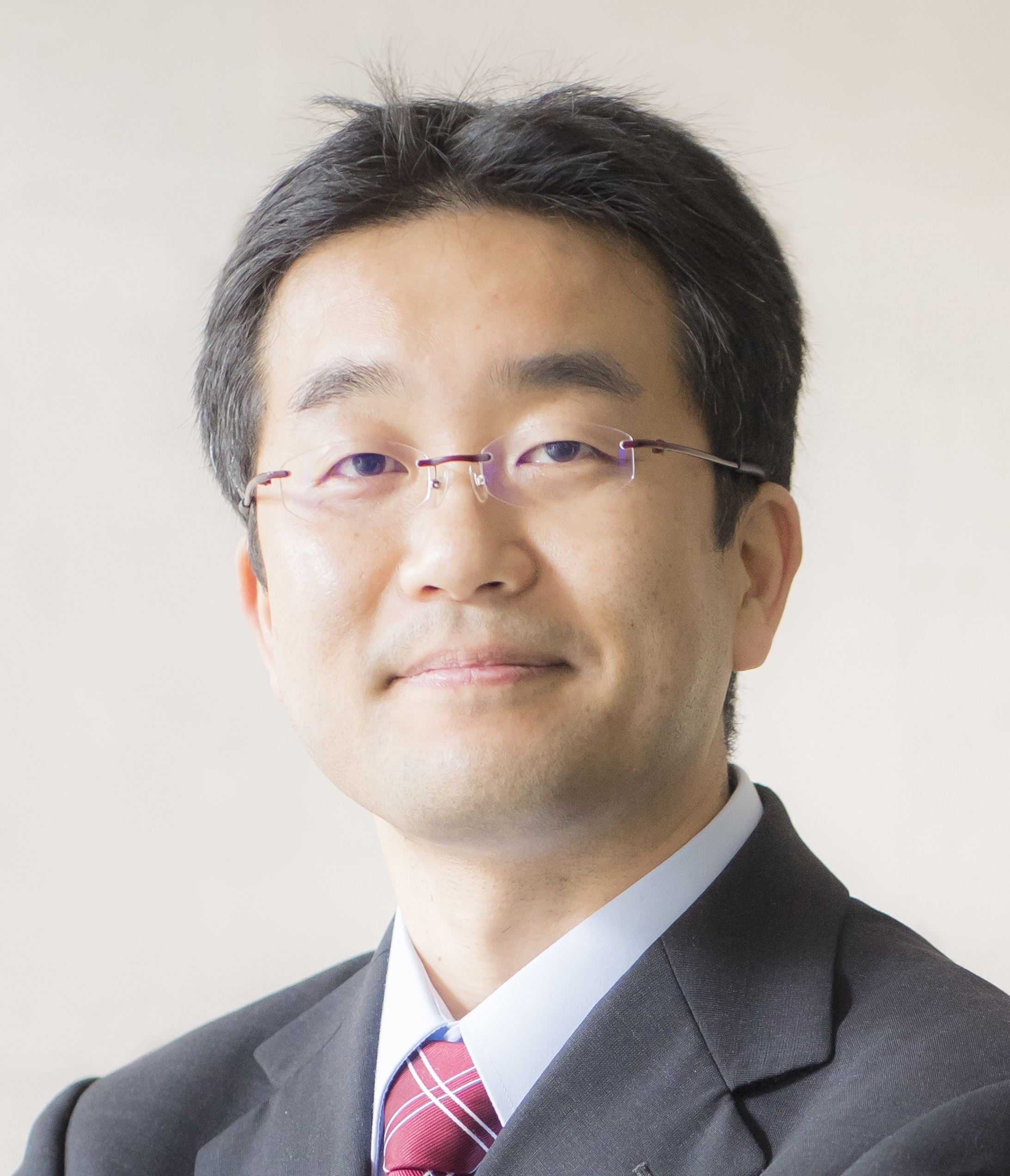}}]{Masashi Sugiyama}
is Director of RIKEN Center for Advanced Intelligence Project and Professor at the University of Tokyo. He received the PhD
degree in computer science from Tokyo Institute of Technology. He served as Program Co-chair for the Neural Information Processing Conference, International Conference on Artificial Intelligence and Statistics, and Asian Conference on Machine Learning. He serves as an Associate Editor for the IEEE Transactions on Pattern Analysis and Machine Intelligence, and an Editorial Board Member for the Machine Learning Journal and Frontiers of Computer Science. He received the Japan Academy Medal in 2017.
\end{IEEEbiography}

\clearpage

\section{Appendix 1: Related Literature}

\subsection{Early Stage}
Before delving into label-noise representation learning, we should briefly overview some of milestone works in label-noise statistical learning. Starting from 1988, Angluin et al. proved that a learning algorithm can handle incorrect training examples robustly, when the noise rate is less than one half under the random noise model~\cite{angluin1988learning}. Bylander further demonstrated that linear threshold functions are polynomially learnable in the presence of classification noise~\cite{bylander1994learning}. Lawrence and Sch\"olkopf constructed a kernel Fisher discriminant to formulate label-noise problem as a probabilistic model~\cite{lawrence2001estimating}, which is solved by Expectation Maximization algorithm. Although the above works explored to tackle noisy labels theoretically and empirically, Bartlett et al. justified that most loss functions are not completely robust to label noise~\cite{bartlett2006convexity}. It means that classifiers based on label-noise robust algorithms are still affected by label noise.

During this period, a lot of works emerged and contributed to this area. For example, Crammer et al. proposed an online Passive-Aggressive perceptron algorithm to cope with label noise~\cite{crammer2006online}. Dredze et al. proposed confidence weighted learning to weigh trusted labels more~\cite{dredze2008confidence}. Freund proposed a boosting algorithm to combat against random label noise~\cite{freund2009more}. To handle label noise theoretically, Cesa-Bianchi et al. proposed an online learning algorithm, leveraging unbiased estimates of the gradient of the loss~\cite{cesa2011online}. Until 2013, Natarajan et al. formally formulated an unbiased risk estimator for binary classification with noisy labels~\cite{natarajan2013learning}. This work is very important to the area, since it is the first work to provide guarantees for risk minimization under random label noise. Moreover, this work provides an easy way to suitably modify any given surrogate loss function for handling label noise.

Meanwhile, Scott et al. studied the classification problem under class-conditional noise model, and propose the way to handle asymmetric label noise~\cite{scott2013classification}. In contrast, van Rooyen et al. proposed the unhinge loss to tackle symmetric label noise~\cite{van2015learning}. Liu and Tao proposed the method of anchor points to estimate the noise rate, and further leverage importance reweighting to design surrogate loss functions for class-conditional label noise~\cite{liu2015classification}. Instead of designing ad-hoc losses, Patrini et al. introduced linear-odd losses, which can be factorized into an even and an odd loss function~\cite{patrini2016loss}. More importantly, they estimated the mean operator from noisy data, and plug this operator in linear-odd losses for empirical risk minimization, which is resistant to asymmetric label noise.

It is noted that, we move from label-noise statistical learning to label-noise representation learning after 2015. There are two reasons behind this phenomenon. First, label-noise statistical learning mainly focus on designing theoretically-robust methods for small-scale noisy data. However, such methods cannot empirically work well on large-scale noisy data in our daily life, such as Clothing1M~\cite{xiao2015learning} emerging from 2015. Second, label-noise statistical learning mainly applies to shallow and convex models, such as support vector machines. However, deep and non-convex models, such as convolutional and recurrent neural networks, have become trendy and mainstream due to the better empirical performance, not only in vision, but also in language, speech and video tasks. Therefore, it is urgent to design label-noise representation learning methods for robustly training of deep models with noisy labels.

\subsection{Emerging Stage}
There are three seminal works in label-noise representation learning with noisy labels. For example, Sukhbaatar et al. introduced an extra but constrained linear ``noise'' layer on top of the softmax layer, which adapts the network outputs to model the noisy label distribution~\cite{sukhbaatar2014training}. Reed et al. augmented the prediction objective with a notion of consistency via a soft and hard bootstrapping~\cite{reed2014training}, where the soft version is equivalent to softmax regression with minimum entropy regularization and the hard version modifies regression targets using the MAP estimation. Intuitively, this bootstrapping procedure provides the learner to disagree with an inconsistent training label, and re-label the training data to improve its label quality. Azadi et al. proposed an auxiliary image regularization technique~\cite{azadi2015auxiliary}. The key idea is to exploit the mutual context information among training data, and encourage the model to select reliable labels.

Followed by seminal works, Goldberger et al. introduced a nonlinear ``noise'' adaptation layer on top of the softmax layer~\cite{goldberger2016training}, which adapts to model the noisy label distribution. Patrini et al. proposed forward and backward loss correction approaches simultaneously~\cite{patrini2017making}. Based on the corrected loss, they explored a robust two-stage training algorithm. A very interesting point is, both Wang et al. and Ren et al. leveraged the same philosophy, namely data reweighting, to learn with label noise. However, they tackled from different perspectives. Specifically, Wang et al. come from a view of Bayesian and propose robust probabilistic modeling~\cite{wang2017robust}, where the posterior of reweighted model will identify uncorrupted data but ignore corrupted data. Ren et al. come from a view of meta-learning~\cite{ren2018learning}, which assigns weights to training samples based on their gradient directions. Namely, their method performs a meta gradient descent step on the current mini-batch example weights (initialized from zero) to minimize the loss on a clean unbiased validation set.

Besides the above works, there are many important works born in 2018, ranging in diverse directions. In high-level, there are several major directions, such as estimating transition matrix, regularization, designing losses and small-loss tricks. Among them, small-loss tricks are inspired by memorization effects of deep neural networks, where deep models will fit easy (clean) patterns first but over-fit hard (noisy) patterns eventually. Namely, small-loss tricks regard small-loss samples as relatively ``clean'' samples, and back-propagate such samples to update the model parameters. For example, Jiang et al. is the first to leverage small-loss tricks to handle label noise~\cite{jiang2018mentornet}. However, they train only a single network iteratively, which is similar to the self-training approach. Such approach inherits the same inferiority of accumulated error caused by the sample-selection bias. To address this issue, Han et al. train two deep neural networks simultaneously, and back propagates the data selected by its peer network and updates itself~\cite{han2018co}.

In the context of representation learning, estimating transition matrix, regularization and designing losses are still prosperous for handling label noise. For instance, given that a small set of trusted examples are available, Hendrycks et al. proposed gold loss correction. Namely, they leveraged trusted examples to estimate the (gold) transition matrix perfectly~\cite{hendrycks2018using}. Therefore, on noisy examples, they will train deep models via forward correction (by gold matrix); on trusted examples, they will train deep models normally. Han et al. proposed ``human-in-the-loop'' idea to easily estimate the transition matrix~\cite{han2018masking}. Specifically, they proposed a human-assisted approach called ``Masking'' that conveys human cognition of invalid class transitions and naturally speculates the structure of the noise transition matrix. Then, they regarded the matrix structure as prior knowledge, which is further incorporated into deep probabilistic modeling.

Moreover, Zhang et al. introduced an implicit regularization called mixup~\cite{zhang2017mixup}, which constructs virtual training data by linear interpolations of features and labels in training data. Mixup encourages the model to behave linearly in-between training examples, which reduces the amount of undesirable oscillations when predicting outside the training examples. Zhang et al. generalized both categorical cross entropy loss and mean absoulte error loss by the negative Box-Cox transformation~\cite{zhang2018generalized}. Their proposed $\mathcal{L}_q$ loss not only has theoretical justification, but also work for both closed-set and open-set noisy labels. Motivated by a dimensionality perspective, Ma [2018] developed a dimensionality-driven learning strategy, which can effectively learn robust low-dimensional subspaces that capture the true data distribution.

\subsection{Flourished Stage}
Starting from 2019, label-noise representation learning has become mature in the top conference venues. Arazo et al. formulated clean and noisy samples as a two-component (clean-noisy) beta mixture model on the loss values~\cite{arazo2019unsupervised}, where the posterior probabilities are then used to implement a dynamically weighted bootstrapping loss. To boost the performance of Co-teaching, Chen et al. introduced the Iterative Noisy Cross-Validation (INCV) method to select a subset of most confident samples (with correct labels)~\cite{chen2019understanding}, while Yu et al. employed the ``Update by Disagreement'' strategy to keep two networks diverged~\cite{yu2019does}. Hendrycks et al. empirically demonstrated that pre-training (i.e., ``pre-train then tune'' paradigm) can improve model robustness against label corruption~\cite{hendrycks2019using}, which is for large-scale noisy datasets.

Under the criteria of balanced error rate (BER) minimization and area under curve (AUC) maximization, Charoenphakdee et al. found that symmetric losses have many merits in combating with noisy labels, even without knowing the noise information. Based on such observation, they proposed Barrier Hinge Loss~\cite{charoenphakdee2019symmetric}. In contrast to selected samples via small-loss tricks, Thulasidasan et al. introduced the abstention-based training, which allows deep abstaining networks to abstain on confusing samples while learning on non-confusing samples~\cite{thulasidasan2019combating}. Following the re-weighting strategy, Shu et al. parameterized the weighting function adaptively as one-layer multilayer perceptron called Meta-Weight-Net~\cite{shu2019meta}, which is free of manually pre-specifying the weighting function.

Entering 2020, Menon et al. mitigated the effects of label noise from an optimization lens, namely using composite loss-based gradient clipping, which naturally introduces the partially Huberised loss for training deep models~\cite{menon2019can}. Nguyen et al. proposed a self-ensemble label filtering method to progressively filter out the wrong labels during training~\cite{nguyen2019self}. Li et al. modeled the per-sample loss distribution with a mixture model to dynamically divide the training data into a labeled set with clean samples and an unlabeled set with noisy samples~\cite{li2020dividemix}. Lyu et al. proposed a provable curriculum loss, which can adaptively select samples for robust stagewise training~\cite{lyu2019curriculum}. Han et al. proposed a versatile approach called scaled stochastic integrated gradient underweighted ascent (SIGUA)~\cite{han2020sigua}. SIGUA uses gradient decent on good data, while using scaled stochastic gradient ascent on bad data rather than dropping those data. After Clothing1M born in 5 years, Jiang et al. proposed a new but realistic type of noisy dataset called ``web-label noise'' (or red noise)~\cite{jiang2020cont}, which enables us to conduct controlled experiments systematically in more realistic scenario.

\section{Appendix 2: Discussion for Three Perspectives}
\subsection{Data Perspective}

It is a typical method to leverage the noise transition matrix for solving the LNRL problem. First, we can insert an adaptation layer into the original network, and this layer can mimic the function of the noise transition matrix. Second, we may keep the original network, but correct the cross-entropy loss via the estimated transition matrix. Lastly, since the accurate matrix estimation will lead to the better classification accuracy, we can use the prior knowledge to ease the estimation burden.

Note that there are other related works from the 
data perspective. For example,
structured noise modeling 
demonstrated that the noise in human-centric annotations exhibits structure, which can be modeled by decoupling the human bias from the correct visually grounded label~\cite{misra2016seeing};
noisy fine-grained recognition showed the potential to train effective models of fine-grained recognition using noisy data from the web only~\cite{krause2016unreasonable};
distillation with side information built a unified distillation framework to use ``side'' information, including a small clean dataset and label relations in a knowledge graph, to combat noisy labels~\cite{li2017learning};
rank pruning addressed the fundamental problem of estimating the noise rates~\cite{northcutt2017learning};
negative learning trained deep networks using complementary labels, which decrease the risk of providing incorrect information~\cite{kim2019nlnl}; 
combinatorial inference reduced the noise level by simply constructing meta-classes and improved the accuracy via combinatorial inferences over multiple constituent classifiers~\cite{seo2019combinatorial}; robust generative adversarial networks (GANs) incorporated a noise transition model, which can learn a clean label conditional generative distribution even when training labels are noisy~\cite{kaneko2019label};
noise-tolerant fairness enabled learning of fair classifiers given noisy sensitive features using the mean-difference score~\cite{lamy2019noise}; and latent class-conditional noise modeled the noise transition in a Bayesian form, namely projecting the noise transition in a Dirichlet-distributed space~\cite{yao2019safeguarded}.

\subsection{Objective Perspective}
Modifying the objective function is another mainstream method to solve the LNRL problem. First, we can augment the objective via either explicit regularizer, e.g., 
Minimum Entropy Regularization~\cite{grandvalet2005semi}, or 
implicit regularizer, e.g., 
Virtual Adversarial Training~\cite{miyato2018virtual}. Second, instead of treating all sub-objective functions equally, we can leverage the reweighting strategy to assign different weights to sub-objective functions. The more weights we assign, the more importance these sub-objective functions have. We can realize the reweighting strategy via different ways, e.g., importance reweighting, a Bayesian method, a mixture model and neural networks. Lastly, we can modify the objective function via redesigning the loss function, e.g., $\ell_q$, barrier hinge loss, partial Huberized loss and curriculum loss. Moreover, we can conduct the label ensemble, e.g., the temporal ensembling and self-ensemble filtering.

Note that there are other related works from the objective perspective. For instance, online crowdsourcing greatly reduces the number of redundant annotations, when crowdsourcing annotations such as bounding boxes, parts, and class labels~\cite{branson2017lean}; an undirected graphical model represents the relationship between noisy and clean labels, where the inference over latent clean labels is tractable and regularized using auxiliary information~\cite{vahdat2017toward}; the active-bias method trains robust deep networks by emphasizing high variance samples~\cite{chang2017active}; model bootstrapped EM jointly models labels and worker quality from noisy crowdsourced data~\cite{khetan2017learning}; the joint optimization framework corrects labels during training by alternating update of network parameters and labels~\cite{tanaka2018joint}; the iterative learning framework trains deep networks with open-set noisy labels~\cite{wang2018iterative}; deep bilevel learning is based on the principles of cross-validation, where a validation set is used to limit the model over-fitting~\cite{jenni2018deep}; symmetric cross entropy (CE) boosts CE symmetrically with a noise robust counterpart,  Reverse Cross Entropy (RCE)~\cite{wang2019symmetric}; the ubiquitous reweighting network learns a robust model from large-scale noisy web data, by considering five key challenges (i.e., imbalanced class sizes, high intra-classes diversity and inter-class similarity, imprecise instances, insufficient representative instances, and ambiguous class labels) in image classification~\cite{li2019learning}; the information-theoretic loss is a generalized version of mutual information, which is provably robust to instance-independent label noise~\cite{xu2019l_dmi}; the peer loss enables learning from noisy labels without requiring a priori specification of the noise rates~\cite{liu2019peer}; and the normalized loss theoretically demonstrates that a simple normalization can make any loss function robust to noisy labels~\cite{ma2020normalized}.

\subsection{Optimization Perspective}
Leveraging memorization effects is an emerging mainstream method to solve the LNRL problem. First, we can combine self-training with memorization effects, which brings us self-paced MentorNet and learning to reweight. Second, we can combine Co-training with memorization effects, which introduces Co-teaching, Co-teaching+, INCV Co-teaching and S2E. Lastly, we can combine Co-training with the SOTA semi-supervised learning MixMatch, which provides DivideMix. Meanwhile, besides memorization, pre-training and deep k-NN are new branches using overparamterized models.

Note that there are other related works from the optimization policy perspective, namely changing training dynamics. For example, multi-task networks jointly learn to clean noisy annotations and accurately classify images~\cite{veit2017learning}; the unified framework of random grouping and attention effectively reduces the negative impact of noisy web image annotations~\cite{zhuang2017attend}; decoupling trains two deep networks simultaneously, and only updates parameters on examples, where there is a disagreement between the two classifiers~\cite{malach2017decoupling};
CleanNet was designed to make label noise
detection and learning from noisy data with human supervision scalable through transfer learning~\cite{lee2018cleannet}; CurriculumNet designs a training curriculum by measuring and ranking the complexity of data using its distribution in a feature space~\cite{guo2018curriculumnet}; Co-mining combines Co-teaching with the Arcface loss~\cite{deng2019arcface} for face recognition tasks~\cite{wang2019co}; O2U-Net only requires adjusting the hyper-parameters of deep networks to make their status transferred from overfitting to underfitting (O2U) cyclically~\cite{huang2019o2u}; deep self-learning is an iterative learning framework
to relabel noisy samples and train deep networks on the real
noisy dataset, without using extra clean supervision~\cite{han2019deep}; the label-noise information strategy is a training method that controls memorization by regularizing label noise information in weights~\cite{harutyunyan2020improving}; different from Co-teaching+, Co-regularization aims to reduce the diversity of two networks during training~\cite{wei2020combating}; and the data coefficient method wisely takes advantage of a small trusted dataset to optimize exemplar weights and labels of mislabeled data, which distills effective supervision for robust training~\cite{zhang2020distilling}.

\section{Appendix 3: Experimental Details}
We conduct experiments on \textit{MNIST} for both Figure~1 and Figure~5. Here, we provide experimental details.

\subsection{Network Structure}
We follow memorization effects~\cite{arpit2017closer} to set up network structure as follows: Input $\rightarrow$ Conv(200,5,5) $\rightarrow$ BN $\rightarrow$ ReLU $\rightarrow$ MaxPool(3,3) $\rightarrow$ Conv(200,5,5) $\rightarrow$ BN $\rightarrow$ ReLU $\rightarrow$ MaxPool(3,3) $\rightarrow$ FC(200,384) $\rightarrow$ BN $\rightarrow$ ReLU $\rightarrow$ FC(384,192) $\rightarrow$ BN $\rightarrow$ ReLU $\rightarrow$ FC(192,10) $\rightarrow$ Softmax.

\subsection{Details of Figure 1}
In Figure~1, we choose \textit{MNIST} with $35\%$ of symmetric noise (Figure~3) as noisy \textit{MNIST}. We choose forward correction (Theorem~2) to correct original $\ell$, which forms corrected $\tilde{\ell}$. Thus, we compare three approaches: 1) $\ell$ on clean \textit{MNIST}; 2) $\tilde{\ell}$ on noisy \textit{MNIST}; and 3) $\ell$ on noisy \textit{MNIST}.

\subsection{Details of Figure 5}
In Figure~5, we choose \textit{MNIST} with $0\%-80\%$ of symmetric noise (Figure~3) as noisy \textit{MNIST} with different noise rates. We choose original $\ell$ to verify memorization effects in deep learning. The training curve will increase gradually. However, the validation curve will increase at the first few epochs, but drop gradually until the convergence.

\end{document}